\documentclass{article}
\usepackage{ijcai21}

\usepackage{times}
\usepackage{soul}
\usepackage{url}
\usepackage[hidelinks]{hyperref}
\usepackage[utf8]{inputenc}
\usepackage[small]{caption}
\usepackage{graphicx}
\usepackage{amsmath}
\usepackage{amsthm}
\usepackage{booktabs}
\usepackage{algorithm}
\usepackage{algorithmic}
\usepackage{subfigure}
\usepackage{titlesec}
\usepackage{enumitem}
\usepackage{amssymb}
\usepackage{amsfonts}       
\usepackage{nicefrac}       
\usepackage{microtype}      
\usepackage{color}
\usepackage{colortbl}
\usepackage{mathrsfs}
\usepackage{listings}
\usepackage{multirow}
\usepackage{wrapfig}
\usepackage{lipsum,booktabs}
\usepackage[algo2e]{algorithm2e} 
\usepackage{footmisc}

\title{Neural Architecture Search of SPD Manifold Networks}

\author{Rhea Sanjay Sukthanker\textsuperscript{\rm 1}, Zhiwu Huang\textsuperscript{\rm 1}, Suryansh Kumar\textsuperscript{\rm 1}, \\ Erik Goron Endsjo\textsuperscript{\rm 1}, Yan Wu\textsuperscript{\rm 1}, Luc Van Gool\textsuperscript{\rm 1,2}\\
    \affiliations
	\textsuperscript{\rm 1}Computer Vision Lab, ETH Z\"urich, Switzerland \quad
	\textsuperscript{\rm 2}PSI, KU Leuven, Belgium\\
	\emails
	rhea.sukthanker@inf.ethz.ch, \{eendsjo, wuyan\}@student.ethz.ch, \\ \{zhiwu.huang, sukumar, vangool\}@vision.ee.ethz.ch
}

\begin{document}
	
	\maketitle
	
	\begin{abstract}
		In this paper, we propose a new neural architecture search (NAS) problem of Symmetric Positive Definite (SPD) manifold networks, aiming to automate the design of SPD neural architectures. 
		To address this problem, we first introduce a geometrically rich and diverse SPD neural architecture search space for an efficient SPD cell design. Further, we model our new NAS problem with a one-shot training process of a single supernet. Based on the supernet modeling, we exploit a differentiable NAS algorithm on our relaxed continuous search space for SPD neural architecture search. Statistical evaluation of our method on drone, action, and emotion recognition tasks mostly provides better results than the state-of-the-art SPD networks and traditional NAS algorithms. Empirical results show that our algorithm excels in discovering better performing SPD network design and provides models that are more than three times lighter than searched by the state-of-the-art NAS algorithms.
	\end{abstract}
	
	\section{Introduction}
	
	Designing an efficient neural network architecture for a given application generally requires a significant amount of time, effort, and domain expertise. To mitigate this issue, there have been emerging a number of neural architecture search (NAS) algorithms to automate the design process of neural architectures \cite{zoph2016neural,liu2017hierarchical,liu2018progressive,liu2018darts,real2019regularized}. However, researchers have barely proposed NAS algorithms to optimize those neural network architecture designs that deal with non-Euclidean data representations and the corresponding set of operations ---to the best of our knowledge.
	
	It is well-known that Symmetric Positive Definite (SPD) manifold-valued data representation has shown overwhelming accomplishments in many real-world applications such as 
	magnetic resonance imaging analysis \cite{pennec2006riemannian}, 
	pedestrian detection \cite{tuzel2008pedestrian}, 
	human action recognition \cite{huang2017riemannian},
	hand gesture recognition \cite{nguyen2019neural}, 
	etc.
	Also, in applications like diffusion tensor imaging of the brain, drone imaging, samples are collected directly as SPD's. As a result, neural network usage based on Euclidean data representation becomes inefficient for those applications. Consequently, this has led to the development of SPD neural networks (SPDNets) \cite{huang2017riemannian} to improve these research areas further. However, the SPDNets are handcrafted, and therefore, the operations or the parameters defined for these networks generally change as per the application. This motivates us to propose a new NAS problem of SPD manifold networks. A solution to this problem can reduce unwanted efforts in SPDNet architecture design.
	Compared to the traditional NAS problem, our NAS problem requires searching for a new basic neural computation cell modeled by a specific directed acyclic graph (DAG), where each node indicates a latent SPD representation, and each edge corresponds to a SPD candidate operation.

	To solve the suggested NAS problem, we exploit a new supernet search strategy that models the architecture search problem as a one-shot training process of a supernet comprised of a mixture of SPD neural architectures. The supernet modeling enables a differential architecture search on a continuous relaxation of SPD neural architecture search space. We evaluate the proposed NAS method on three benchmark datasets, showing the automatically searched SPD neural architectures perform better than the state-of-the-art handcrafted SPDNets for radar, facial emotion, and skeletal action recognition. In summary, our work makes the following contributions:
	\begin{itemize}[leftmargin=*,topsep=0pt, noitemsep]
		
		\item We introduce a brand-new NAS problem of SPD manifold networks that opens up a novel research problem at the intersection of automated machine learning and SPD manifold learning. For this problem, we exploit a new search space and a new supernet modeling, both of which respect the particular Riemannian geometry of SPD manifold networks.
		
		\item We propose a sparsemax-based NAS technique to optimize sparsely mixed architecture candidates during the search stage. This reduces the discrepancy between the search on mixed candidates and the training on one single candidate. To optimize the new NAS objective, we exploit a new bi-level algorithm with manifold- and convex-based updates.

		\item Evaluation on three benchmark datasets shows that our searched SPD neural architectures can outperform handcrafted SPDNets \cite{huang2017riemannian,brooks2019riemannian,chakraborty2020manifoldnet} and the state-of-the-art NAS methods \cite{liu2018darts,chu2019fair}. Our searched architecture is more than three times lighter than those searched using traditional NAS algorithms.
		\vspace{-0.1cm}
	\end{itemize}

	\section{Background} {\label{ss:background}}
	
	As our work is directed towards solving a new NAS problem, we confine our discussion to the work that has greatly influenced our method \emph{i.e.}, one-shot NAS methods and SPD networks.
	There are mainly two types of one-shot NAS methods based on the architecture modeling \cite{elsken2018neural}  \emph{(i) parameterized architecture} \cite{liu2018darts,zheng2019multinomial,wu2019fbnet,chu2019fair}, and  \emph{(ii) sampled architecture} \cite{deb2002fast,chu2019scarletnas}. In this paper, we adhere to the parametric modeling due to its promising results on conventional neural architectures. A majority of the previous work on NAS with continuous search space fine-tunes the explicit feature of specific architectures  \cite{saxena2016convolutional,veniat2018learning,ahmed2017connectivity,shin2018differentiable}. On the contrary, \cite{liu2018darts,liang2019darts+,zhou2019bayesnas,DSO-NAS,wu2020neural,chu2019fair} provides architectural diversity for NAS with highly competitive performances. The other part of our work focuses on SPD network architectures. There exist algorithms to develop handcrafted SPDNets like \cite{huang2017riemannian,brooks2019riemannian,chakraborty2020manifoldnet}. To automate the process of SPD network design, in this work, we choose the most promising approaches from the field of NAS \cite{liu2018darts} and SPD networks \cite{huang2017riemannian}). 
	
	Next, we summarize some of the essential notions of Riemannian geometry of SPD manifolds, followed by introducing some basic SPDNet operations and layers. As some of the introduced operations and layers have been well-studied by the existing literature, we apply them directly to define our SPD neural architectures' search space.
	
	\smallskip
	\noindent
	\textbf{Representation and Operation.}\label{ssec:spdgeom}
	We denote $n\times n$ real SPD as $\boldsymbol{X}\in \mathcal{S}^n_{++}$. A real SPD matrix $\boldsymbol{X}\in\mathcal{S}^n_{++}$ satisfies the property that for any non-zero $z \in \mathbb{R}^n$, $z^T \boldsymbol{X} z>0$. We denote $\mathcal{T}_{\boldsymbol{X}} \mathcal{M}$  as the tangent space of the manifold $\mathcal{M}$ at $\boldsymbol{X} \in \mathcal{S}^n_{++}$ and $
	\log$ corresponds to matrix logarithm.  
	
	Let $\boldsymbol{X}_1, \boldsymbol{X}_2$ be any two points on the SPD manifold then the distance between them is given by $ \delta_{\mathcal{M}}(\boldsymbol{X}_1, \boldsymbol{X}_2)  = 0.5\|\log(\boldsymbol{X}_1^{-\frac{1}{2}} \boldsymbol{X}_2  \boldsymbol{X}_1^{-\frac{1}{2}})\|_{F}$.
	There are other efficient methods to compute distance between two points on the SPD manifold, however, their discussion is beyond the scope of our work \cite{gao2019robust,dong2017deep}. Other property of the Riemannian manifold of our interest is local diffeomorphism of geodesics which is a one-to-one mapping from the point on the tangent space of the manifold to the manifold \cite{pennec2020manifold,Lackenbyriemannian}. To define such notions, let $\boldsymbol{X} \in \mathcal{S}_{++}^n$ be the base point and, $\boldsymbol{Y} \in \mathcal{T}_{\boldsymbol{X}}\mathcal{S}_{++}^n$, then  $\boldsymbol{Y} \in \mathcal{T}_{\boldsymbol{X}}\mathcal{S}_{++}^n$ is associated to a point on the SPD manifold \cite{pennec2020manifold} by the map
	\begin{equation}\label{eq:expmap}
	\begin{small}
	\begin{aligned}
	\exp_{\boldsymbol{X}} (\boldsymbol{Y}) = \boldsymbol{X}^{\frac{1}{2}} \exp(\boldsymbol{X}^{-\frac{1}{2}} \boldsymbol{Y}  \boldsymbol{X}^{-\frac{1}{2}}) \boldsymbol{X}^{\frac{1}{2}} \in \mathcal{S}_{++}^n, ~~\forall ~\boldsymbol{Y} \in \mathcal{T}_{\boldsymbol{X}}.
	\end{aligned}
	\end{small}
	\end{equation}
	Likewise, 
	\begin{equation}
	\begin{small}
	\begin{aligned}
	\log_{\boldsymbol{X}}(\boldsymbol{Z}) = \boldsymbol{X}^{\frac{1}{2}} \log(\boldsymbol{X}^{-\frac{1}{2}} \boldsymbol{Z}  \boldsymbol{X}^{-\frac{1}{2}}) \boldsymbol{X}^{\frac{1}{2}} \in \mathcal{T}_{\boldsymbol{X}} \end{aligned} \end{small}, ~\forall ~\boldsymbol{Z} \in \mathcal{S}_{++}^n
	\end{equation}
	is defined as its inverse map.
	
	\noindent
	\textbf{1) Basic operations of SPD Network.}
	Operations such as mean centralization, normalization, and adding bias to a batch of data are inherent performance booster for neural networks. Accordingly, works like \cite{brooks2019riemannian,chakraborty2020manifoldnorm} use the notion of such operations for the manifold-valued data to define analogous operations on manifolds. Below we introduce them following \cite{brooks2019riemannian} work.
	\smallskip
	\begin{itemize}[leftmargin=*,topsep=0pt]
		\item \emph{Batch mean, centering and bias}: Given a batch of $N$ SPD matrices $\small \{\boldsymbol{X}_{i}\}_{i=1}^{N}$, we can compute its Riemannian barycenter ($\mathscr{B}$) as:
		\begin{equation}
		\mathscr{B} = \underset{\boldsymbol{X}_{\mu} \in \mathcal{S}_{++}^n} {\textrm{argmin}}
		\sum_{i=1}^{N} 
		\delta_{\mathcal{M}}^2 (\boldsymbol{X}_{i}, \boldsymbol{X}_{\mu}).
		\end{equation}
		
		It is sometimes referred to as Fr\'echet mean \cite{moakher2005differential,bhatia2006riemannian}. This definition can be extended to compute the weighted Riemannian Barycenter also known as weighted Fr\'echet Mean (wFM)\footnote{Following \cite{tuzel2008pedestrian,brooks2019riemannian}, we focus on the estimate of wFM using Karcher flow.}
		\begin{equation}\label{eq:batchmean}
		\begin{small}
		\begin{aligned}
		\mathscr{B} = \underset{\boldsymbol{X}_{\mu} \in \mathcal{S}_{++}^n}{\textrm{argmin}} \sum_{i=1}^{N} w_i \delta_{\mathcal{M}}^2 (\boldsymbol{X}_{i}, \boldsymbol{X}_{\mu}); ~~\text{s.t.} ~~w_i \geq 0, ~\sum_{i=1}^N w_i = 1.
		\end{aligned}
		\end{small}
		\end{equation}
		Eq:(\ref{eq:batchmean}) can be estimated using Karcher flow \cite{karcher1977riemannian,bonnabel2013stochastic,brooks2019riemannian} or recursive geodesic mean \cite{cheng2016recursive,chakraborty2020manifoldnet}.  
	\end{itemize}

	\begin{figure*}[t]
		\centering
        \includegraphics[width=0.92\linewidth]{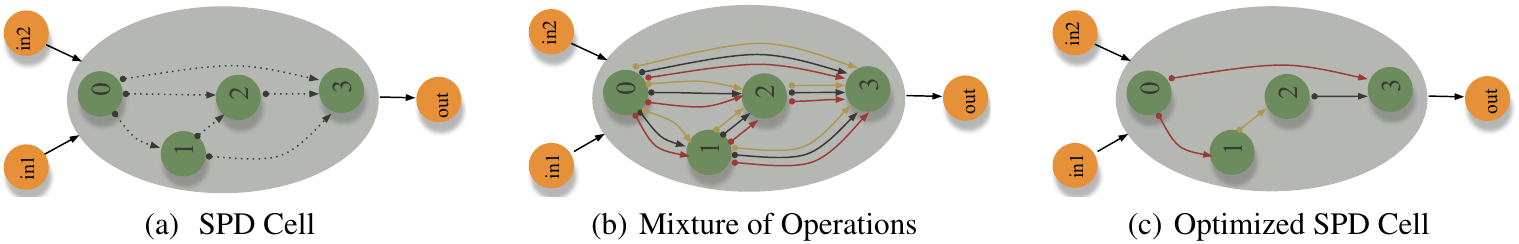}
        
		\vspace{-0.3cm}
		\caption{\small (a) A SPD cell structure composed of 4 SPD nodes, 2 input node and 1 output node. Initially the edges are unknown (b) Mixture of candidate SPD operations between nodes (c) Optimal cell architecture obtained after solving the bi-level optimization formulation.}
		\label{fig:cellintuition}
	\end{figure*}
	
	\smallskip
	\noindent
	\textbf{2) Basic layers of SPD Network.}  Analogous to standard convolutional networks (ConvNets),  \cite{huang2017riemannian,brooks2019riemannian,chakraborty2020manifoldnet} designed SPD layers to perform operations that respect SPD manifold constraints. Assuming $\boldsymbol{X}_{k-1} \in \mathcal{S}^{n}_{++}$ be the input SPD matix to the $k^{th}$ layer, the SPD layers are defined as follows:
	
	\begin{itemize}[leftmargin=*, topsep=0pt]
		\item \emph{ BiMap layer}: This layer corresponds to a dense layer for SPD data. It reduces the dimension of a input SPD matrix via a transformation matrix $\boldsymbol{W}_{k}$ as
		$\boldsymbol{X}_{k} = \boldsymbol{W}_{k} \boldsymbol{X}_{k-1} \boldsymbol{W}_{k}^{T}$. To ensure $\boldsymbol{X}_{k}$ to be SPD, $\boldsymbol{W}_{k}$ is commonly required to be of full row-rank through an orthogonality constraint on it.
		
		\item \emph{Batch normalization layer:} To perform batch normalization after each BiMap layer, we first compute the Riemannian barycenter of one batch of SPD samples followed by a running mean update step, which is Riemannian weighted average between the batch mean and the current running mean, with the weights $(1 - \theta)$ and $(\theta)$ respectively. With the mean, we centralize and add a bias to each SPD sample in the batch using Eq:(\ref{eq:SPDcentering}), where $\mathscr{P}$ is the notation used for parallel transport and $I$ is the identity matrix:
		\begin{equation}\label{eq:SPDcentering}
		\begin{small}
		\begin{aligned}
		\displaystyle & \textrm{Centering the} ~\mathscr{B}: \boldsymbol{X}_i^{c} =  \mathscr{P}_{\mathscr{B} \rightarrow I} (\boldsymbol{X}_i)  = \mathscr{B}^{-\frac{1}{2}} \boldsymbol{X}_i \mathscr{B}^{-\frac{1}{2}}, \\
		\displaystyle & 
		\textrm{Bias towards} ~\boldsymbol{G}: \boldsymbol{X}_i^{b} =  \mathscr{P}_{I \rightarrow \boldsymbol{G}} (\boldsymbol{X}_i^c)  = \boldsymbol{G}^{\frac{1}{2}} \boldsymbol{X}_i^c \boldsymbol{G}^{\frac{1}{2}}.
		\end{aligned}
		\end{small}
		\end{equation}
		
		\item \emph{ReEig layer}: The ReEig layer is analogous to ReLU layers presented in the classical ConvNets. It aims to introduce non-linearity to SPD networks. The ReEig for the $k^{th}$ layer is defined as:
		$
		\boldsymbol{X}_{k}$ = $\boldsymbol{U}_{k-1} \max (\epsilon \boldsymbol{I}, \Sigma_{k-1}) \boldsymbol{U}_{k-1}^{T}$ $~\textrm{where}, \boldsymbol{X}_{k-1}$ = $\boldsymbol{U}_{k-1} \Sigma_{k-1} \boldsymbol{U}_{k-1}^{T}
		$, $\boldsymbol{I}$ is the identity matrix, and $\epsilon>0$ is a rectification threshold value. $U_{k-1}, \Sigma_{k-1}$ are the orthonormal matrix and singular-value matrix respectively, and obtained via matrix factorization of $\boldsymbol{X}_{k-1}$.
		
		\item \emph{LogEig layer}: To map the manifold representation of SPDs to a flat space so that a Euclidean operation can be performed, LogEig layer is introduced. The LogEig layer is defined as:
		$\boldsymbol{X}_{k} = \boldsymbol{U}_{k-1} \log (\Sigma_{k-1}) \boldsymbol{U}_{k-1}^{T} ~\textrm{where}, \boldsymbol{X}_{k-1} = \boldsymbol{U}_{k-1} \Sigma_{k-1} \boldsymbol{U}_{k-1}^{T}$. The LogEig layer is used with fully connected layers to solve tasks with SPD representation.
		
		\item \emph{ExpEig layer}: The ExpEig layer is designed to map the corresponding SPD representation from a flat space back to an SPD manifold space. It is defined as $
		\boldsymbol{X}_{k} = \boldsymbol{U}_{k-1} \exp (\Sigma_{k-1}) \boldsymbol{U}_{k-1}^{T}, ~\textrm{where} \quad \boldsymbol{X}_{k-1} = \boldsymbol{U}_{k-1} \Sigma_{k-1} \boldsymbol{U}_{k-1}^{T}$.
		
		\item \emph{Weighted Riemannian pooling layer}: It uses wFM to perform the weighted pooling on SPD samples.
		To compute wFM,
		we follow \cite{brooks2019riemannian} to use Karcher flow \cite{karcher1977riemannian} due to its simplicity and wide usage.
		
	\end{itemize}

	\begin{table*}[t]
		
		\small
		\centering
		\begin{tabular}{|c|c|c|c|}
			\hline
			\rowcolor[gray]{0.85}
			\textbf{Operation} & \textbf{Definition}  & \textbf{Operation}  & \textbf{Definition} \\ \hline
			BiMap\_0     & \{BiMap, Batch Normalization\}        & WeightedReimannPooling\_normal  & \{wFM on SPD multiple times\}  \\ \hline
			BiMap\_1     & \{BiMap,Batch Normalization, ReEig\}  & AveragePooling\_reduced  & \{LogEig, AveragePooling, ExpEig\}  \\ \hline
			BiMap\_2     & \{ReEig, BiMap, Batch Normalization\} & MaxPooling\_reduced  & \{LogEig, MaxPooling, ExpEig\} \\ \hline
			Skip\_normal & \{Output same as input\}              & \multicolumn{2}{c|}{\multirow{2}{*}{\begin{tabular}[c]{@{}c@{}}Skip\_reduced = \{$\text{C}_\text{in}$ = BiMap($\text{X}_\text{in}$), $[\text{U}_\text{in}, \text{D}_\text{in}, \sim] = \textrm{svd}(\text{C}_\text{in}); \text{in = 1, 2}$\}, \\ $\text{C}_\text{out} = \text{U}_\text{b} \text{D}_\text{b} \text{U}_\text{b}^{T}$, \text{where}, $\text{U}_\text{b} = \text{diag}(\text{U}_1, \text{U}_2) ~\text{and} ~\text{D}_\text{b} = \textrm{diag}(\text{D}_1, \text{D}_2)$\end{tabular}}} \\ \cline{1-2}
			None\_normal & \{Return identity matrix\}            & \multicolumn{2}{c|}{}                                                      \\ \hline
		\end{tabular}
		\caption{Search space for the proposed SPD architecture search method. }\label{tab:cell_operations}
	\end{table*}
	
	\section{Proposed Method}\label{ss:problem}
	
	Firstly, we introduce a new definition of the computation cell for our task. In contrast to traditional NAS computation cell design  \cite{liu2018darts,chu2019fair}, our computational cell (called SPD cell), incorporates the notion of SPD manifold geometry while performing any SPD operations. Similar to the basic NAS cell design, our SPD cell can either be a \textbf{normal cell} that returns SPD feature maps of the same width and height or, a \textbf{reduction cell} in which the SPD feature maps are reduced by a certain factor in width and height. Secondly,  solving our new NAS problem will require an appropriate and diverse SPD \emph{search space} that can help NAS method to optimize for an effective SPD cell, which can then be stacked and trained to build an efficient SPD neural network architecture.
	
	Concretely, an SPD cell is modeled by a directed asyclic graph (DAG) which is composed of nodes and edges. In our DAG each \emph{node} indicates an \emph{latent representation} of the SPD manifold-valued data \emph{i.e.}, an intermediate SPD feature map, and each \emph{edge} corresponds to a \emph{valid candidate operation} on the SPD manifold (see Fig.\ref{fig:cellintuition}(a)). Each edge of a SPD cell is associated with a set of candidate SPD manifold operations ($\mathscr{O}_{\mathcal{M}}$) that transforms the SPD-valued latent representation from the source node  (say $\boldsymbol{X}_{\mathcal{M}}^{(i)}$) to the target node  (say $ \boldsymbol{X}_{\mathcal{M}}^{(j)}$). We define the intermediate transformation between the nodes in our SPD cell as:
	\begin{equation}
	\boldsymbol{X}_{\mathcal{M}}^{(j)} =  \underset{\boldsymbol{X}_{\mathcal{M}}^{(j)} } {\textrm{argmin}}  \sum_{i<j} \delta_{\mathcal{M}}^2 \Big(\mathscr{O}_{\mathcal{M}}^{(i, j)}\big(\boldsymbol{X}_{\mathcal{M}}^{(i)}\big), \boldsymbol{X}_{\mathcal{M}}^{(j)}\Big),
	\end{equation}
	where $\delta_{\mathcal{M}}$ denotes the geodesic distance. This transformation result corresponds to the unweighted Fr\'echet mean of the operations based on the predecessors, such that the mixture of all operations still reside on SPD manifolds. Note that our definition of SPD cell ensures that each computational graph preserves the appropriate geometric structure of the SPD manifold. Equipped with the notion of SPD cell and its intermediate transformation, we are ready to propose our search space (\S \ref{ss:oursearchspace}) followed by the solution to our new NAS problem (\S \ref{ss:superNetmethod}) and its results (\S \ref{sec:experimentsandresults}).

	\subsection{Search Space}\label{ss:oursearchspace}
	Our search space consists of a set of valid SPD network operations.
	It includes some existing SPD operations, \emph{e.g.}, BiMap, Batch Normalization, ReEig, LogEig, ExpEig, and weighted Riemannian pooling layers, all of which are introduced in Sec.\ref{ss:background}. Though existing works have well explored those individual operations (e.g., BiMap, LogEig, ExpEig), their different combinations are still understudied and are essential to enrich our search space. To enhance the search space, following traditional NAS methods \cite{liu2018darts,gong2019autogan}, we apply the SPD batch normalization to every SPD convolution operation (\emph{i.e.}, BiMap), and design three variants of convolution blocks including the one without activation (\emph{i.e.}, ReEig), the one using post-activation and the one using pre-activation (see Table \ref{tab:cell_operations}). 
	In addition, we introduce \textbf{five new operations} analogous to \cite{liu2018darts} to enrich the search space in the context of SPD networks. These are, skip normal, none normal, average pooling, max pooling, and skip reduced. The effect of such diverse operation choices has not been fully explored for SPD networks. All the candidate operations are illustrated in Table (\ref{tab:cell_operations}), and their definitions are as follows:

	\textbf{(a) Skip normal}: It preserves the input representation and is similar to regular skip connections. \textbf{(b) None normal}: It corresponds to the operation that returns identity as the output i.e, the notion of zero in the SPD space.
	\textbf{(c) Max pooling}: Given a set of SPD matrices, the max pooling operation first projects these samples to a flat space via a LogEig operation, where  a standard max pooling operation is performed. Finally, an ExpEig operation is used to map the samples back to the SPD manifold. \textbf{(d) Average pooling}: Similar to Max pooling, the average pooling operation first projects the samples to the flat space using a LogEig operation, where a standard average pooling is employed. To map the sample back to the SPD manifold, an ExpEig operation is used. \textbf{(e) Skip reduced}: It is similar to `skip\_normal' but in contrast, it decomposes the input into small matrices to reduces the inter-dependency between channels. Our definition of the reduce operation is in line with the work of \cite{liu2018darts}.
	
	The newly introduced operations allow us to generate a more diverse discrete search space. As presented in Table (\ref{tab:comparison_NATO_HDM05}), the randomly selected architecture (generally consisting of the newly introduced SPD operations) shows some improvement over the handcrafted SPDNets,  which only contain conventional SPD operations. This establishes the effectiveness of the introduced rich search space. For more details about the proposed SPD operations, please refer to our Appendix.
	
	\subsection{Supernet Search} \label{ss:superNetmethod}
	To solve the new NAS problem, one of the most promising methodologies is supernet modeling.
	In general, the supernet method models the architecture search problem as a one-shot training process of a single supernet that consists of all candidate architectures. Based on the {supernet modeling}, we search for the optimal SPD neural architecture by parameterizing the design of the supernet architectures, which is based on the {continuous relaxation of the SPD neural architecture representation}. Such an approach allows for an efficient search of architecture using the gradient descent approach. Next, we introduce our supernet search method, followed by a solution to our proposed bi-level optimization problem. Fig.\ref{fig:cellintuition}(b) and Fig.\ref{fig:cellintuition}(c) illustrate an overview of our proposed method.

\begin{figure*}
		\centering
		\subfigure[\label{fig:activation_weight}]{\includegraphics[width=0.55\linewidth]{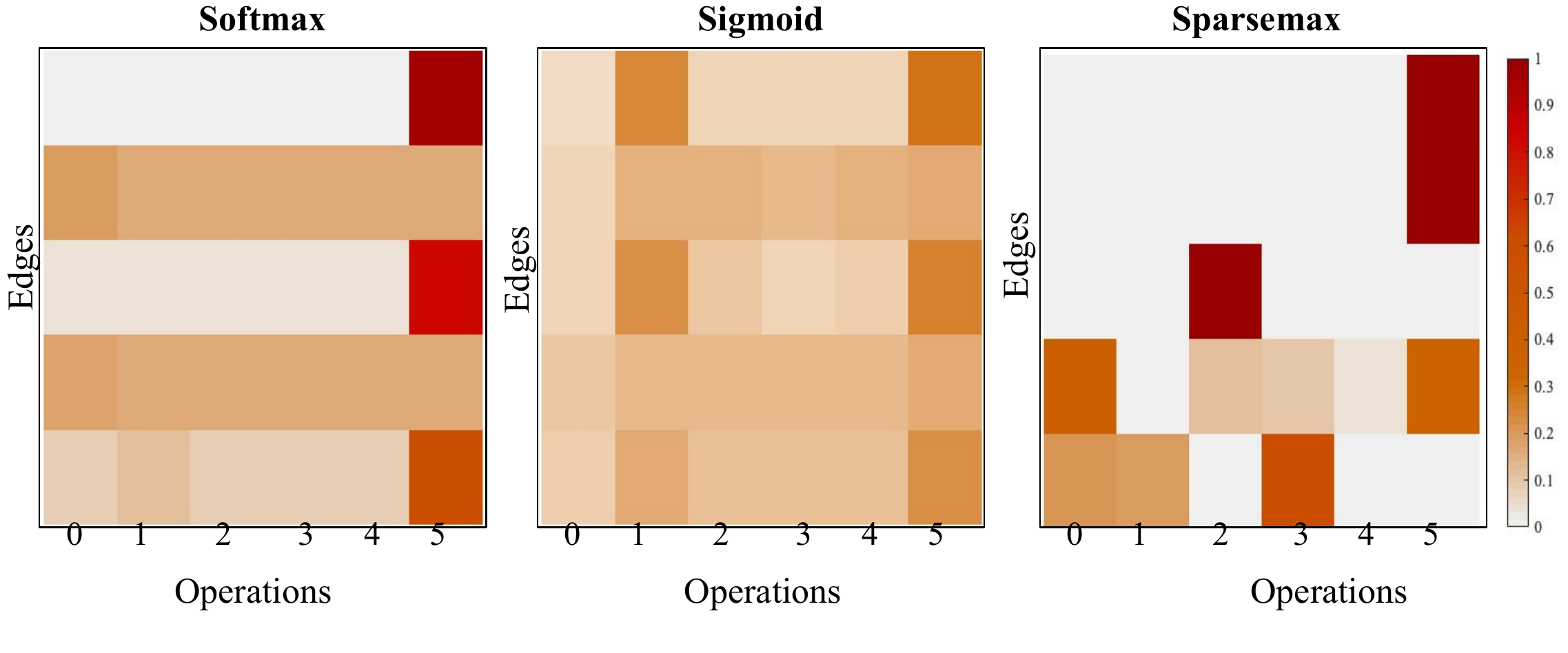}}
		~~\subfigure[\label{fig:RADAR_HDMO5_sparsegenotypes}]{\includegraphics[width=0.42\linewidth]{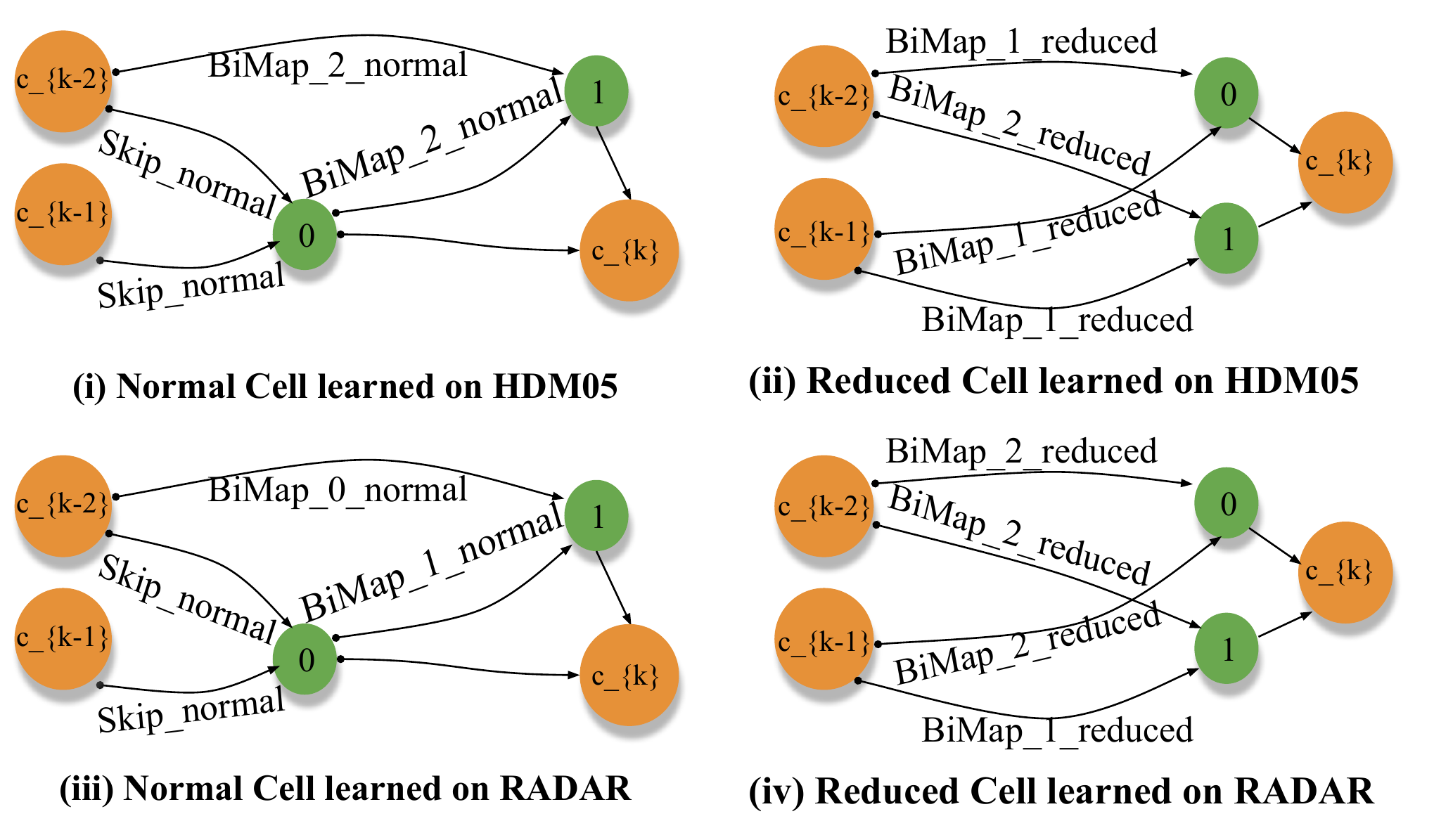}}
		\vspace{-0.5cm}
		\caption{\small(a) Distribution of edge weights for operation selection using softmax, sigmoid, and sparsemax on the Fr\'echet mixture of SPD operations. (b) Derived sparsemax architecture by the proposed SPDNetNAS. The better sparsity leads to less skips and poolings compared to those of other NAS solutions shown in our Appendix. }
		\label{fig:stats}
		\vspace{-0.4cm}
	\end{figure*}

	To search for an optimal SPD architecture parameterized by $\alpha$, we optimize the over parameterized supernet. In essence, it stacks the basic computation cells with the parameterized candidate operations from our search space in a one-shot search manner. 
	The contribution of specific subnets to the supernet helps in deriving the optimal architecture from the supernet.
	Since the proposed operation search space is discrete in nature, we relax the explicit choice of an operation to make the search space continuous. To do so, we use wFM over all possible candidate operations. Mathematically, 
	\begin{equation}\label{eq:mixtureofoperation}
	\begin{small}
	\begin{aligned}
	& \Bar{\mathscr{O}}_{\mathcal{M}}(\boldsymbol{X}_{\mathcal{M}})  = \underset{\boldsymbol{X}_{\mathcal{M}}^{\mu} } {\textrm{argmin}}  \sum_{k=1}^{N_e} \Tilde{\alpha}^k \delta_{\mathcal{M}}^2 \Big(\mathscr{O}_{\mathcal{M}}^{(k)}\big(\boldsymbol{X}_{\mathcal{M}}\big), \boldsymbol{X}_{\mathcal{M}}^{\mu}\Big); \\ & ~\textrm{subject to:}  ~\mathbf{1}^T\Tilde{\alpha}=1, ~0 \leq \Tilde{\alpha} \leq 1,
	\end{aligned}
	\end{small}
	\end{equation}
	where $\mathscr{O}_{\mathcal{M}}^k$ is the $k^{th}$ candidate operation between nodes, $\boldsymbol{X}_\mu$ is the intermediate SPD manifold mean (Eq.\ref{eq:batchmean}) and, $N_e$ denotes number of edges.  We can compute wFM solution either using Karcher flow \cite{karcher1977riemannian} or recursive geodesic mean \cite{chakraborty2020manifoldnet} algorithm. Nonetheless, we adhere to Karcher flow algorithm as it is widely used to calculate wFM. To impose the explicit convex constraint on $\Tilde{\alpha}$, we project the solution onto the probability simplex as
	\begin{equation}\label{eq:simplexconstriant}
	\begin{small}
	\begin{aligned}
	\underset{\alpha} {\textrm{minimize}} ~\|\alpha -\Tilde{\alpha}\|_2^2; ~\textrm{subject to:} ~~\mathbf{1}^T\alpha=1, ~0 \leq \alpha \leq 1.
	\end{aligned}
	\end{small}
	\end{equation}
	Eq:(\ref{eq:simplexconstriant}) enforces the explicit constraint on the weights to supply $\alpha$ for our task and can easily be added as a convex layer in the framework \cite{agrawal2019differentiable}. This projection is likely to reach the boundary of the simplex, in which case $\alpha$ becomes sparse \cite{martins2016softmax}.
	Optionally, softmax, sigmoid and other regularization methods can be employed to satisfy the convex constraint. However,  \cite{chu2019fair} has  observed that the use of softmax can cause performance collapse and may lead to aggregation of skip connections. While \cite{chu2019fair} suggested sigmoid can overcome the unfairness problem with softmax, it may output smoothly changed values which is hard to threshold for dropping redundant operations with non-marginal contributions to the supernet. Also, the regularization in \cite{chu2019fair}, may not preserve the summation equal to 1 constraint.  Besides, \cite{chakraborty2020manifoldnet} proposes recursive statistical approach to solve wFM with convex constraint, however, the definition proposed do not explicitly preserve the equality constraint and it requires re-normalization of the solution. In contrast, our approach composes of the sparsemax transformation for convex Fr\'echet mixture of SPD operations with the following two advantages: 1) It can preserve most of the important properties of softmax such as, it is simple to evaluate, cheaper to differentiate \cite{martins2016softmax}. 2) It is able to produce \textbf{sparse distributions} such that the best operation associated with each edge is likely to make more dominant contributions to the supernet, and thus better architectures can be derived (refer to Fig~\ref{fig:activation_weight},\ref{fig:RADAR_HDMO5_sparsegenotypes} and \S \ref{sec:experimentsandresults}).

	From Eq:(\ref{eq:mixtureofoperation}--\ref{eq:simplexconstriant}), the \textbf{mixing of operations} between nodes is determined by the weighted combination of alpha's ($\alpha^{k}$) and the set of operations ($\mathscr{O}_{\mathcal{M}}^k$). This relaxation makes the search space continuous and therefore, architecture search can be achieved by learning a set of alpha ($\alpha = \{\alpha^k, ~\forall ~k ~\in N_e\}$). To achieve our goal, we simultaneously learn the contributions (i.e., the architecture parameterization) $\alpha$ of all the candidate operations and their corresponding network weights ($w$). Consequently, for a given $w$, we optimize $\alpha$ and vice-versa resulting in the following bi-level optimization problem.
	\begin{equation}\label{eq:bilevelopt}
	\centering
	\small
	\begin{aligned}
	& \displaystyle    \underset{\alpha}{\textrm{min.}} ~\boldsymbol{E}_{val}^{U} \big(w^{opt}(\alpha), \alpha\big); ~\textrm{s.t.} ~w^{opt}(\alpha) = \underset{w}{\textrm{argmin}} ~\boldsymbol{E}_{train}^{L} (w, \alpha),
	\end{aligned}
	\end{equation}
	where the lower-level optimization ($\boldsymbol{E}_{train}^{L}$) corresponds to the optimal weight variable learned for a given $\alpha$ \emph{i.e.}, $w^{opt}(\alpha)$ using a training loss. The upper-level optimization ($\boldsymbol{E}_{val}^{U}$) solves for the variable $\alpha$ given the optimal $w$ using a validation loss.
	This bi-level search method gives an optimal mixture of multiple small architectures. To derive each node in the discrete architecture, we maintain top-$k$ operations \emph{i.e.}, the ${k}^\text{{th}}$ highest weight among all the candidate operations associated with all the previous nodes.
	
	\paragraph{Bi-level Optimization.} The bi-level optimization problem proposed in Eq:(\ref{eq:bilevelopt}) is difficult to solve. 
	On one hand, the $\alpha$ can be interpreted as hyper-parameter but it's not a scalar and hence, harder to optimize. On the other hand, the lower optimization is computationally expensive.
	Following \cite{liu2018darts} work,  we approximate $w^{opt}(\alpha)$ in the upper- optimization problem to skip inner-optimization as follows:
	\begin{equation}\label{eq:bilevelapprox}
	\begin{small}
	\begin{aligned}
	\nabla_\alpha\boldsymbol{E}_{val}^{U} \big(w^{opt}(\alpha), \alpha\big)
	\approx \nabla_\alpha \boldsymbol{E}_{val}^{U} \big(w- \eta \nabla_w\boldsymbol{E}_{train}^{L}(w, \alpha), \alpha\big),
	\end{aligned}
	\end{small}
	\end{equation}
	Here, $\eta$ is the learning rate and $\nabla$ is the gradient operator. Note that the gradient based optimization for $w$ must follow the geometry of SPD manifolds to update the structured connection weight, and its corresponding SPD matrix data. Applying the chain rule to Eq:(\ref{eq:bilevelapprox}) gives
	\begin{equation}\label{eq:chainruleapprox}
	\centering
	\small
	\overbrace{\nabla_\alpha \boldsymbol{E}_{val}^{U} \big(\Tilde{w}, \alpha\big)}^{\text{first term}}
	- \overbrace{\eta \nabla_{\alpha, w}^2 \boldsymbol{E}_{train}^{L}(w, \alpha)
		\nabla_{\Tilde{w}}\boldsymbol{E}_{val}^{U} (\Tilde{w}, \alpha)}^{\text{second term}},
	\end{equation}
	where, $\Tilde{w} = \mathbf{\Psi_{r}}\big(w - \eta \Tilde{\nabla}_w \boldsymbol{E}_{train}^{L}(w, \alpha)\big)$ denotes the weight update on the SPD manifold for the forward model. $\Tilde{\nabla}_w$, $\mathbf{\Psi_r}$ symbolizes the Riemannian gradient and the retraction operator respectively. The second term in the Eq:(\ref{eq:chainruleapprox}) involves second order differentials with very high computational complexity, hence, using the finite approximation method  the second term of Eq:(\ref{eq:chainruleapprox}) reduces to:
	\begin{equation}\label{eq:finiteorder}
	\centering
	\small
	\frac{\nabla_\alpha \boldsymbol{E}_{train}^{L} (w^+, \alpha) - \nabla_\alpha \boldsymbol{E}_{train}^{L} (w^-, \alpha)}{2 \delta}
	\end{equation}
	where, ${w}^{\pm} = \mathbf{\Psi_{r}}(w \pm \delta \Tilde{\nabla}_{\Tilde{w}}\boldsymbol{E}_{val}^{U}(\Tilde{w}, \alpha))$ and $\delta$ is a small number set to $0.01/\| \nabla_{\Tilde{w}}\boldsymbol{E}_{val}^{U} (\Tilde{w}, \alpha) \|_2$.
	
	Though the optimization follows the suggested structure in \cite{liu2018darts}, there are some key differences. Firstly, the updates on the manifold-valued kernel weights are constrained on manifolds, which ensures that the feature maps at every intermediate layer are SPDs. For concrete derivations on back-propagation for SPD network layers, refer to \cite{huang2017riemannian} work. Secondly, the update on the aggregation weights of the involved SPD operations needs to satisfy an additional strict convex constraint, which is enforced as a part of the optimization problem Eq:(\ref{eq:simplexconstriant}). The pseudo code of our method is outlined in \textbf{Algorithm} \ref{algo:1}.

	\begin{algorithm}[H]
		\SetAlgoLined
		\While{not converged}{
			\textbf{Step1}: Update the architecture parameter $\alpha$ using Eq:(\ref{eq:bilevelopt}) solution by satisfying an additional strict convex constraint. Note that updates on those manifold parameter $w$  of the network follow the gradient descent on SPD manifolds;\\
			\textbf{Step2}: Update $w$ by solving $\nabla_w\boldsymbol{E}_{train}(w, \alpha)$; Enforce the SPD manifold gradients to update those structured $w$  \cite{huang2017riemannian,brooks2019riemannian};
		} \textbf{Ensure}: Final architecture based on $\alpha$. Decide the operation at an edge $k$ using $\underset{o \in \mathscr{O}_{\mathcal{M}} }{\text{argmax}} \{\alpha_{o}^k\}$
		\caption{\small The proposed SPDNetNAS}\label{algo:1}

	\end{algorithm}

	\begin{table*}[h]
		
		\vspace{-0.2cm}
		\scriptsize
		\centering
		\begin{tabular}{c|cc|cc|cc}
			\hline
			\rowcolor[gray]{0.6}
			Dataset & DARTS & FairDARTS  & SPDNet  & SPDNetBN  & SPDNetNAS (R) & SPDNetNAS  \\ \hline
			RADAR & 98.21\%$\pm$ 0.23 & \textbf{98.51\% $\pm$ 0.09} & 93.21\% $\pm$ 0.39   & 92.13\% $\pm$ 0.77 & 95.49\% $\pm$ 0.08 & 97.75\% $\pm$ 0.30 \\ \hline
			
			\rowcolor[gray]{0.85}
			\#RADAR &  2.6383 MB & 2.6614 MB   & 0.0014 MB & 0.0018 MB & 0.0185 MB & 0.0184 MB \\ \hline
			
			
			HDM05 & 54.27\% $\pm$ 0.92 & 58.29\% $\pm$ 0.86  & 61.60\% $\pm$ 1.35  & 65.20\% $\pm$ 1.15  & 66.92\% $\pm$ 0.72  &   \textbf{69.87\% $\pm$ 0.31} \\ \hline 
			
			\rowcolor[gray]{0.85}
			\#HDM05 & 3.6800MB & 5.1353 MB& 0.1082 MB  &  0.1091 MB  & 1.0557 MB  &  1.064MB MB \\ \hline 
			
			
		\end{tabular}
	 	\vspace{-0.2cm}
	\caption{Performance comparison of our SPDNetNAS against existing SPDNets and traditional NAS on drone and action recognition. SPDNetNAS (R): randomly select architecure from our search space, DARTS/FairDARTS: accepts logarithm forms of SPDs. The search time of our method on RADAR and HDM05 is noted to be 1 CPU days and 3 CPU days respectively. And the search cost of DARTS and FairDARTS on RADAR and HDM05 are about 8 GPU hours. \#RADAR and \#HDM05 show the model sizes on the respective dataset. }\label{tab:comparison_NATO_HDM05}
	\end{table*}
	
	\begin{table*}[h]
		
		\scriptsize
		\centering
		\begin{tabular}{cc|ccc|cc}
			\hline
			\rowcolor[gray]{0.85}
			DARTS & FairDARTS & ManifoldNet & SPDNet  & SPDNetBN & SPDNetNAS (RADAR) & SPDNetNAS (HDM05) \\ \hline
			26.88 \% & 22.31\% & 28.84\% & 34.06\% & 37.80\% & \textbf{40.80\%}  & \textbf{40.64}\% \\ \hline
		\end{tabular}
	 \vspace{-0.2cm}
	\caption{Performance comparison of our SPDNetNAS transferred architectures on AFEW against handcrafted SPDNets and Euclidean NAS. SPDNetNAS(RADAR/HDM05): architectures searched on RADAR and HDM05 respectively. }\label{tab:comparison_afew}
		\vspace{-0.3cm}
	\end{table*}

	\section{Experiments and Results}\label{sec:experimentsandresults}
	
	{To keep the experimental evaluation consistent with the existing SPD networks \cite{huang2017riemannian,brooks2019riemannian}, we follow them to use RADAR \cite{chen2006micro}, HDM05 \cite{muller2007documentation}, and AFEW \cite{dhall2014emotion} datasets. }
	For our SPDNetNAS\footnote{{Source code link: \url{https://github.com/rheasukthanker/spdnetnas}}.}, we first optimize the supernet on the training/validation sets and then prune it with the best operation for each edge. Finally, we train the optimized architecture from scratch to document the results. For both these stages, we consider the same normal and reduction cells. A cell receives preprocessed inputs using fixed BiMap\_2 to make the input of the same initial dimension. All architectures are trained with a batch size of 30. Learning rate ($\eta$) for RADAR, HDM05, and AFEW  dataset is set to 0.025, 0.025 and 0.05 respectively. Besides, we conducted experiments where we select architecture using a random search path (SPDNetNAS (R)) to justify whether our search space with the introduced SPD operations can derive meaningful architectures.

	We compare SPDNet \cite{huang2017riemannian}, SPDNetBN \cite{brooks2019riemannian}, and ManifoldNet \cite{chakraborty2020manifoldnet} that are handcrafted SPD networks. SPDNet and SPDNetBN are evaluated using their original implementations and default setup, and ManifoldNet is evaluated following the video classification setup of \cite{chakraborty2020manifoldnet}.
	It is non-trivial to adapt ManifoldNet to RADAR and HDM05, as ManifoldNet requires SPD features with multiple channels, and both of the two datasets can hardly obtain them. For comparing against Euclidean NAS methods, we used DARTS \cite{liu2018darts}, and FairDARTS \cite{chu2019fair} by treating SPD's logarithm maps\footnote{Feeding raw SPDs generally results in performance degradation.} as Euclidean data in their official implementation with the default setup.

	\noindent
	\textbf{a) Drone Recognition.} For this task, we use the RADAR dataset from \cite{chen2006micro}. This dataset's synthetic setting is composed of radar signals, where each signal is split into windows of length 20, resulting in a 20x20 covariance matrix for each window (one radar data point). The synthesized dataset consists of 1000 data points per class. Given $20 \times 20$ input covariance matrices, our reduction cell reduces them to $10 \times 10$ matrices followed by the normal cell to provide a higher complexity to our network. Following \cite{brooks2019riemannian}, we assign 50\%, 25\%, and 25\% of the dataset for training, validation, and test set, respectively. The Euclidean NAS algorithms are evaluated on the Euclidean map of the input. For direct SPD input, DARTS performance (95.86\%) and FairDarts (92.26\%) are worse as expected. For this dataset, our algorithm takes 1 CPU day of search time to provide the SPD architecture. Training and validation take 9 CPU hours for 200 epochs. Test results on this dataset are provided in Table~(\ref{tab:comparison_NATO_HDM05}), which clearly shows our method's benefit. The statistical performance shows that our NAS method provides an architecture with much fewer parameters (more than 140 times) than well-known Euclidean NAS on this task. The normal and reduction cells obtained on this dataset are shown in Fig.~\ref{fig:RADAR_HDMO5_sparsegenotypes}. {The rich search space of our algorithm allows inclusion of skips and poolings (unlike traditional SPDNets) in our architectures thus improving the performance.}

	\noindent
	\textbf{b) Action Recognition.} For this task, we use the HDM05 dataset \cite{muller2007documentation} which contains 130 action classes, yet, for consistency with previous work \cite{brooks2019riemannian}, we used 117 class for performance comparison. This dataset has 3D coordinates of 31 joints per frame. Following the earlier works  \cite{harandi2017dimensionality}, we model action for a sequence using $93 \times 93$ joint covariance matrix. The dataset has 2083 SPD matrices distributed among all 117 classes. Similar to the previous task, we split the dataset into 50\%, 25\%, and 25\% for training, validation, and testing. Here, our reduction cell is designed to reduce the matrices dimensions from 93 to 30 for legitimate comparison against \cite{brooks2019riemannian}. To search for the best architecture, we ran our algorithm for 50 epoch (3 CPU days). Figure \ref{fig:RADAR_HDMO5_sparsegenotypes} shows the final cell architecture that got selected based on the validation performance. The optimal architecture is trained from scratch for 100 epochs, which took approximately 16 CPU hours. The test accuracy achieved on this dataset is provided in Table~(\ref{tab:comparison_NATO_HDM05}). Statistics clearly show that our models perform better despite being lighter than the NAS models and the handcrafted SPDNets. The Euclidean NAS models' inferior results show the use of SPD geometric constraint for SPD valued data.

	\noindent
	\textbf{c) Emotion Recognition.} We use the AFEW dataset \cite{dhall2014emotion} to evaluate the transferability of our searched architecture for emotion recognition. This dataset has 1345 videos of facial expressions classified into 7 distinct classes. To train on the video frames directly, we stack all the handcrafted SPDNets, and our searched SPDNet on top of a convolutional network \cite{meng2019frame} with its official implementation. 
	For ManifoldNet, we compute a $64\times64$ spatial covariance matrix for each frame on the intermediate ConvNet features of $64\times56\times56$ (channels, height, width). We follow the reported setup of \cite{chakraborty2020manifoldnet} to first apply a single wFM layer with kernel size 5, stride 3, and 8 channels, followed by three temporal wFM layers of kernel size 3 and stride 2, with the channels being 1, 4, 8 respectively. We closely follow the official implementation of ManifoldNet for the wFM layers and adapt the code to our specific task. Since SPDNet, SPDNetBN, and our SPDNetNAS require a single channel SPD matrix as input, we use the final 512-dimensional vector extracted from the convolutional network, project it using a dense layer to a 100-dimensional feature vector and compute a $100\times100$ temporal covariance matrix. 
	To study our algorithm's transferability, we evaluate its searched architecture on RADAR and HDM05. Also, we evaluate DARTS and FairDARTS directly on the video frames of AFEW. Table (\ref{tab:comparison_afew}) reports the evaluations results. As we can observe, the transferred architectures can handle the new dataset quite convincingly, and their test accuracies are better than those of the state-of-the-art SPDNets and Euclidean NAS algorithms. In Appendix\textsuperscript{\ref{note1}}, we present the results of these competing methods and our searched models on the raw SPD features of the AFEW dataset.

	\noindent
\textbf{d) Comparison under similar model complexities.}
We compare the statistical performance of our method against the other competing methods under similar model sizes. Table (\ref{tab:comparison_size}) shows the results obtained on the RADAR dataset. One key point to note here is that when we increase the number of parameters in SPDNet and SPDNetBN,  we observe a very severe degradation in the performance accuracy ---mainly because the network starts overfitting rapidly.  The performance degradation is far more severe for the HDM05 dataset with SPDNet (1.047MB) performing 0.7619\% and SPDNetBN (1.082MB) performing 1.45\% and hence,  is not reported in Table (\ref{tab:comparison_size}). That further indicates the ability of SPDNetNAS to generalize better and avoid overfitting despite the larger model size.

\begin{table}[t]
	\vspace{-0.2cm}
	\scriptsize
	\centering
	\begin{tabular}{c|c|c|c|c}
		\hline
		\rowcolor[gray]{0.6}
		Dataset &Manifoldnet & SPDNet  & SPDNetBN   & SPDNetNAS  \\ \hline
		RADAR & NA
		&73.066\%
		&  87.866\%
		& 
		\textbf{97.75}\%
		\\ \hline
		
		\rowcolor[gray]{0.85}
		\#RADAR & NA &  0.01838 MB
		&  0.01838 MB  & 0.01840 MB \\ \hline
		AFEW &25.8\%
		&  
		34.06\%
		& 
		37.80\%
		& \textbf{40.64}\%
		\\ \hline
		
		\rowcolor[gray]{0.85}
		\#AFEW & 11.6476 MB &  11.2626 MB
		&  11.2651 MB  & 11.7601 MB \\ \hline
		
	\end{tabular}
		\vspace{-0.2cm}
	\caption{Performance of our SPDNetNAS against existing SPDNets with comparable model sizes on RADAR and AFEW.}\label{tab:comparison_size}
	\vspace{-0.2cm}
\end{table}

	\noindent
\textbf{e) Ablation study.} 
Lastly, we conducted an ablation study to realize the effect of probability simplex constraint (sparsemax) on our suggested Fr\'echet mixture of SPD operations. Although in Fig. \ref{fig:activation_weight} we show better probability weight distribution with sparsemax, Table(\ref{tab:activation_comparison}) shows that it performs better empirically as well on both RADAR and HDM05 compared to the softmax and sigmoid cases. Therefore, SPD architectures derived using the sparsemax is observed to be better.
\begin{table}[t]
	\scriptsize
	\centering
	\vspace{-3pt}
	\begin{tabular}{c|c|c|c}
		\hline
		\rowcolor[gray]{0.85}
		Dataset & softmax  & sigmoid & sparsemax \\ \hline
		RADAR  & 96.47\% $\pm$ 0.10  & 97.70\% $\pm$ 0.23 & \textbf{97.75\% $\pm$ 0.30}\\ \hline
		HDM05  & 68.74\% $\pm$ 0.93 & 68.64\% $\pm$ 0.09  & \textbf{69.87\% $\pm$ 0.31}  \\ \hline
	\end{tabular}
		\vspace{-0.2cm}
	\caption{Ablations study on different solutions to our suggested Fr\'echet mixture of SPD operations within SPDNetNAS.}
\label{tab:activation_comparison}
	\vspace{-0.3cm}
\end{table}

	\vspace{-0.1cm}
	\section{Conclusion and Future Direction}
	
	We presented a neural architecture search problem of SPD manifold networks. To address it, we proposed a new differentiable NAS algorithm that consists of sparsemax-based Fr\'echet relaxation of search space and manifold update-based bi-level optimization. Evaluation on several benchmark datasets showed the clear superiority of the proposed NAS method over handcrafted SPD networks and Euclidean NAS algorithms. 
	
	As a future work, it would be interesting to generalize our NAS algorithm to cope with other manifold valued data (e.g., \cite{huang2017deep,huang2018building,chakraborty2018statistical,kumar2018scalable,zhen2019dilated,kumar2019jumping,kumar2020dense}) and manifold poolings (e.g., \cite{wang2017g2denet,engin2018deepkspd,wang2019deep}), which are generally valuable for visual recognition, structure from motion, medical imaging, radar imaging, forensics, appearance tracking to name a few.  
	
	\section*{Acknowledgments}
	This work was supported in part by the ETH Z\"urich Fund (OK), an Amazon AWS grant, and an Nvidia GPU grant. Suryansh Kumar’s project is supported by ``ETH Z\"urich Foundation and Google, Project Number: 2019-HE-323'' for bringing together best academic and industrial research. The authors would like to thank Siwei Zhang from ETH Z\"urich for evaluating our code on AWS GPUs.
	
	\bibliographystyle{named}
	\bibliography{ijcai21_short}
	
\appendix	
\section*{Technical Appendix}

In the appendix, we will introduce some more details about the proposed SPD operations and the suggested optimization. Finally, we present additional experimental analysis for our proposed method.

\section{Detailed Description of Our Proposed SPD Operations}

In this section, we describe some of the major operations defined in the main paper 
from an intuitive point of view. We particularly focus on some of the new operations that are defined for the input SPDs, i.e., the Weighted Riemannian Pooling, the Average/Max Pooling, the Skip Reduced operation and the Mixture of Operations.

\subsection{Weighted Riemannian Pooling}
\begin{figure}[!htbp]
\centering
{\includegraphics[width=1\linewidth]{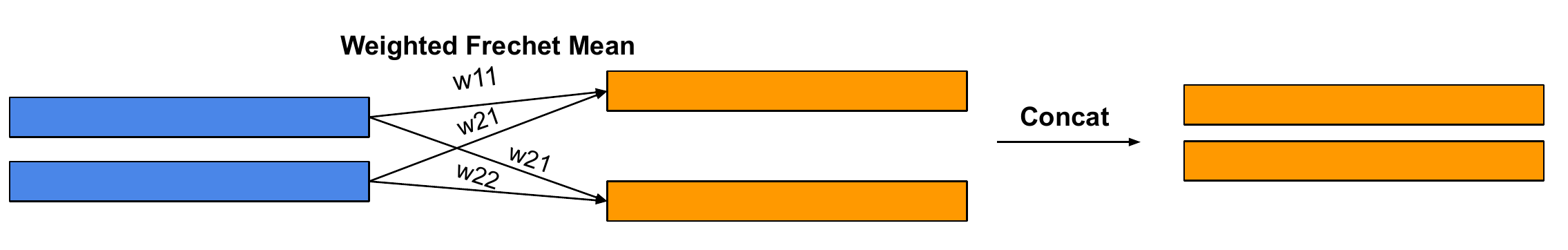}}
\caption{\small Weighted Riemannian Pooling: Performs multiple weighted Fr\'echet means on the channels of the input SPD }
\label{fig:weightedpooling}
\end{figure}

Figure~\ref{fig:weightedpooling} provides an intuition behind the Weighted Riemannian Pooling
operation. Here, w\_11, w\_21, etc., corresponds to the set of normalized weights for each channel (shown as two blue channels). The next channel ---shown in orange, is then computed as weighted Fr\'echet mean over these two input channels.  This procedure is repeated to achieve the desired number of output channels (here two), and finally all the output channels are concatenated. The weights are learnt as a part of the optimization procedure ensuring the explicit convex constraint is imposed. 

\subsection{Average and Max Pooling}
\begin{figure}[!htbp]
\centering
{\includegraphics[width=1\linewidth]{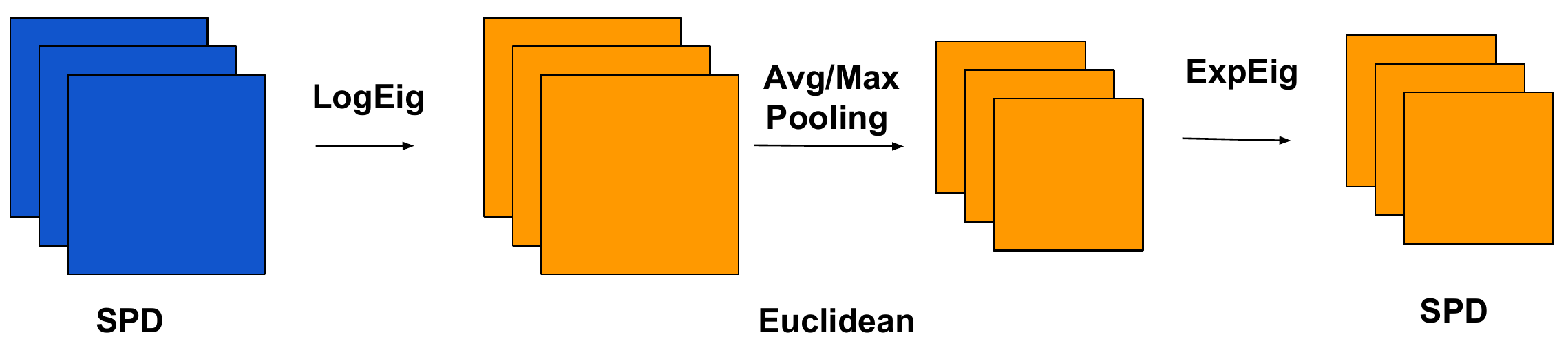}}
\caption{\small Avg/Max Pooling: Maps the SPD matrix to Euclidean space using LogEig mapping, does avg/max pooling followed by ExpEig map}
\label{fig:avgmaxpooling}
\end{figure}

In Figure~\ref{fig:avgmaxpooling} we show our average and max pooling operations. We first perform a LogEig map on the SPD matrices to project them to the Euclidean space. Next, we perform average and max pooling on these Euclidean matrices similar to classical convolutional neural networks. We further perform an ExpEig map to project the Euclidean matrices back on the SPD manifold. The diagram shown in Figure~\ref{fig:avgmaxpooling} is inspired by \cite{huang2017riemannian} work. The kernel size of \textbf{AveragePooling\_reduced} and \textbf{MaxPooling\_reduced} is set to 2 or 4 for all experiments according to the specific dimensionality reduction factors.

\subsection{Skip Reduced}
\begin{figure}[h]
\centering
{\includegraphics[width=1\linewidth]{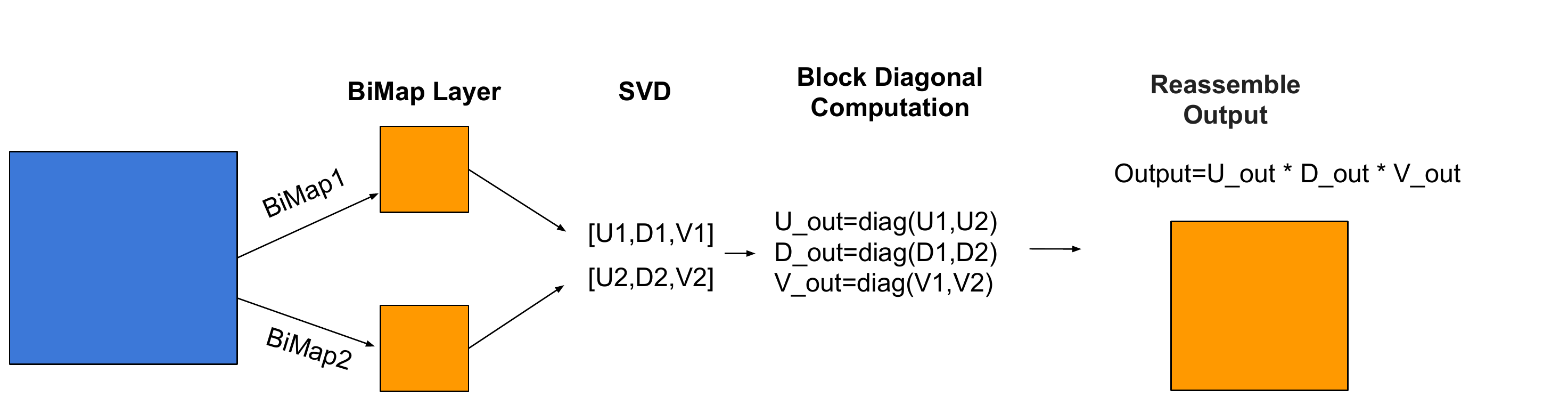}}
\caption{\small Skip Reduced:Maps input to two smaller matrices using BiMaps, followed by SVD decomposition on them and then computes the output using a block diagonal form of U's D's and V's }
\label{fig:skipreduced}
\end{figure}

Following \cite{liu2018darts}, we defined an analogous of Skip operation on a single channel for the reduced cell (Figure~\ref{fig:skipreduced}). We start by using a BiMap layer ---equivalent to Conv in \cite{liu2018darts}, to map the input channel to an SPD whose space dimension is half of the input dimension. We further perform an SVD decomposition on the two SPDs followed by concatenating the Us, Vs and Ds obtained from SVD to block diagonal matrices. Finally, we compute the output by multiplying the block diagonal U, V and D computed before. 

\subsection{Mixed Operation on SPDs}
\begin{figure}[h]
\centering

{\includegraphics[width=1\linewidth]{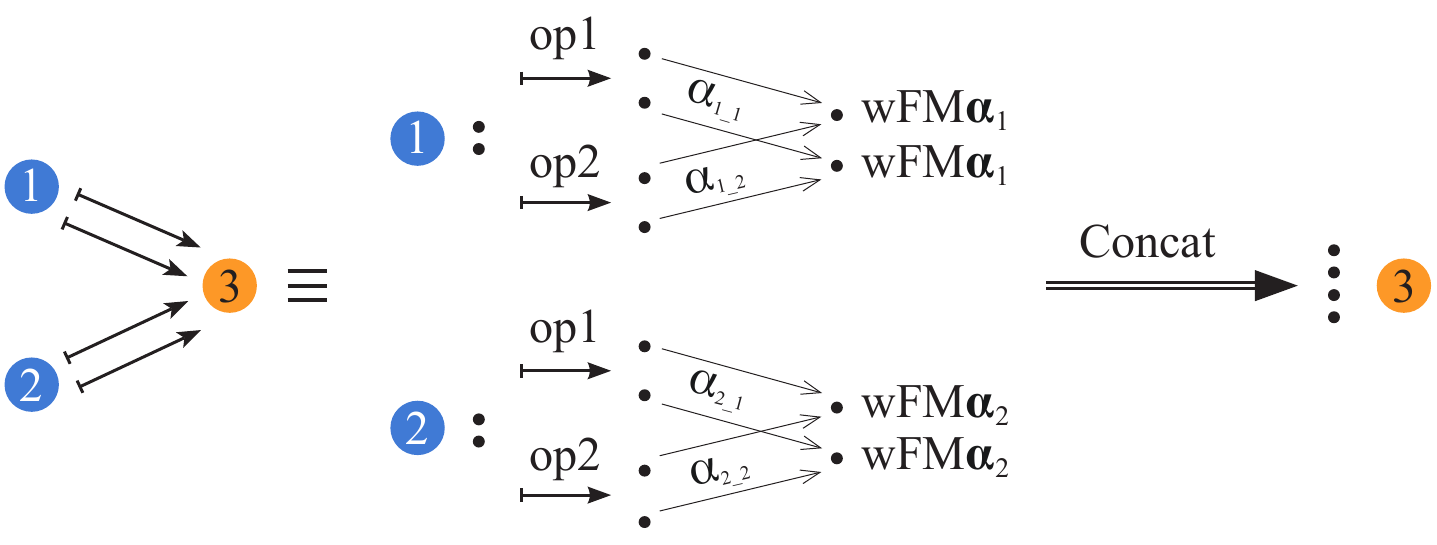}}
\caption{\small Detailed overview of mixed operations. We simplify the example by taking 3 nodes (two input nodes and one output node) and two candidate operations. Input nodes have two channels (SPD matrices), we perform channelwise weighted Fréchet mean between the result of each operation (edge) where weights $\alpha$'s are optimized during bi-level architecture search optimization. Output node 3 is formed by concatenating both mixed operation outputs, resulting in a four channel node. }
\label{fig:mixop}
\end{figure}

In Figure~\ref{fig:mixop} we provide an intuition of the mixed operation we have proposed in the main paper. We consider a very simple base case of three nodes, two input nodes (1 and 2) and one output node (node 3). The goal is to compute the output node 3 from input nodes 1 and 2. We perform a candidate set of operations on the input node, which correspond to edges between the nodes (here two for simplicity). Each operation has a weight $\alpha_{i\_j}$ where i corresponds to the node index and j is the candidate operation identifier. In Figure~\ref{fig:mixop} below i and j $\in \{1,2\}$ and $ \boldsymbol{\alpha_{1}}=\{\alpha_{1\_1},\alpha_{1\_2}\}$ , $ \boldsymbol{\alpha_{2}}=\{\alpha_{2\_1},\alpha_{2\_2}\}$ . $\alpha$'s are optimized as a part of the bi-level optimization procedure proposed in the main paper. Using these alpha's, we perform a channel-wise weighted Fr\'echet mean (wFM) as depicted in the figure below. This effectively corresponds to a mixture of the candidate operations. Note that the alpha's corresponding to all channels of a single operation are assumed to be the same. Once the weighted Fr\'echet means have been computed for nodes 1 and 2, we perform a channel-wise concatenation on the outputs of the two nodes, effectively doubling the number of channels in node 3. 

\section{Suggested Optimization Details}

\begin{figure}[h]
\centering
{\includegraphics[width=1\linewidth]{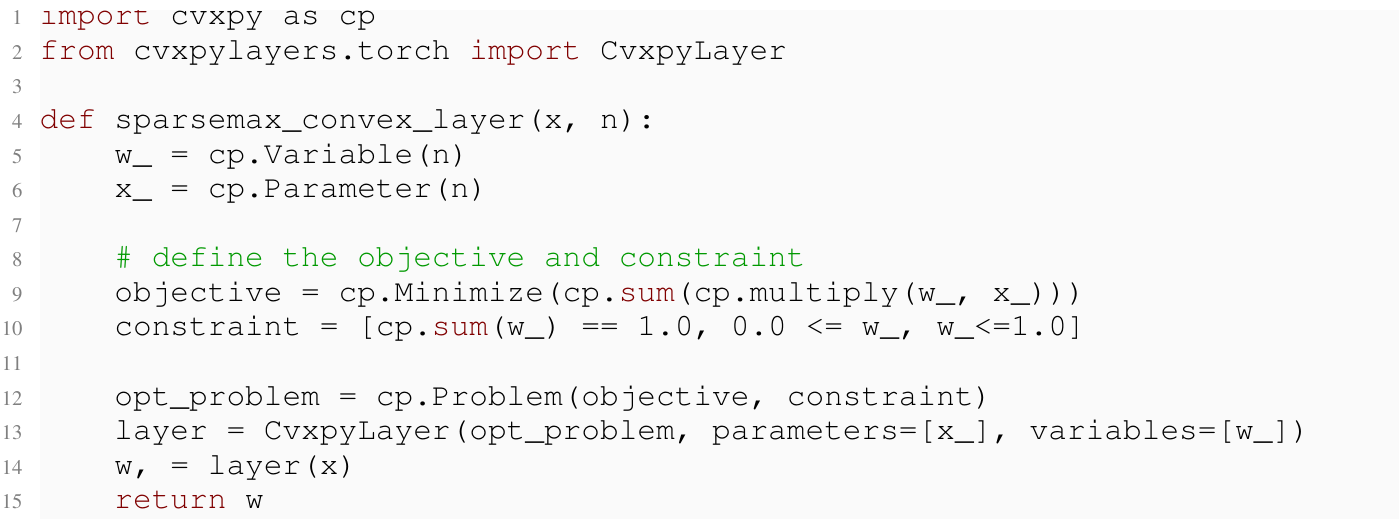}}
\caption{\small Function to Solve the sparsemax constraint optimization}
\label{fig:convexlayer}
\end{figure}

As described in the paper we suggest to replace the standard softmax constraint commonly used in differentiable architecture search with the sparsemax constraint. In addition to ensuring that it outputs valid probabilities (summing to 1), sparsemax is capable of producing sparse distribution, i.e., preserving only a few dominant contributions. This further helps in discovery of a more efficient and optimal architecture.\\
Figure \ref{fig:convexlayer} shows the implementation of the differentiable convex layer for our suggested sparsemax optimization. We use a convex optimization layer to explicitly ensure that the $w_i>0$ and $\sum_i w_i=1$ constraints for a valid Fréchet mean on the SPD manifold are enforced.

    
    
    

\section{Additional Experimental Analysis}






\subsection{Effect of modifying preprocessing layers for multiple dimensionality reduction}
Unlike \cite{huang2017riemannian} work on the SPD network, where multiple transformation matrices are applied at multiple layers to reduce the dimension of the input data, our reduction cell presented in the main paper is one step. For example: For HDM05 dataset \cite{muller2007documentation}, the author's of SPDNet \cite{huang2017riemannian} apply  $93 \times 70$, $70 \times 50$,  $50 \times 30$, transformation matrices to reduce the dimension of the input matrix, on the contrary, we reduce the dimension in one step from 93 to 30 which is inline with \cite{brooks2019riemannian} work. 

\begin{figure*}[!htbp]
\centering
    \subfigure[\label{fig:normalmultidim} Normal cell]{\includegraphics[width=0.56\linewidth]{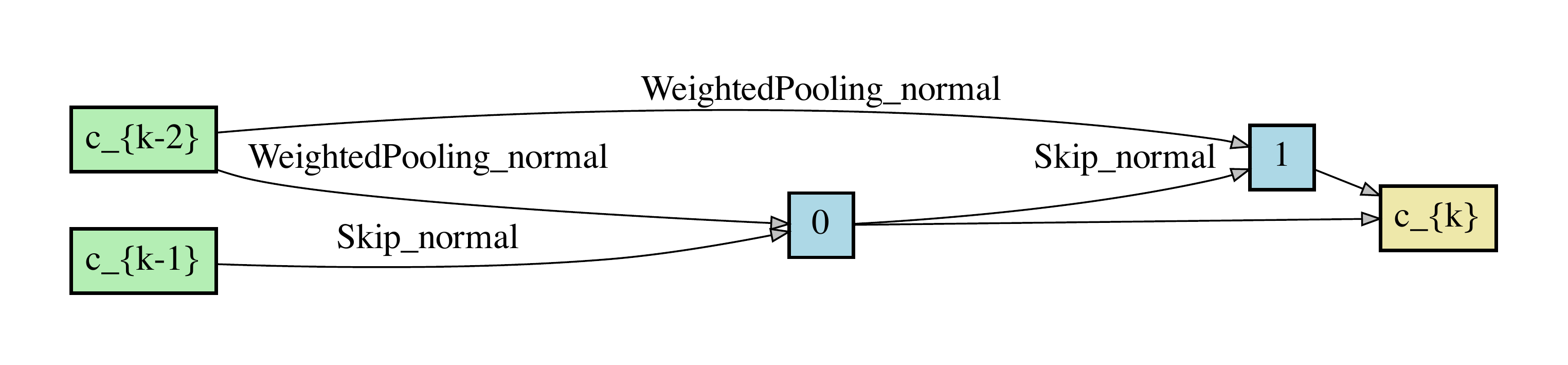}}
    \subfigure[\label{fig:reducemultidim} Reduction cell]{\includegraphics[width=0.4\linewidth]{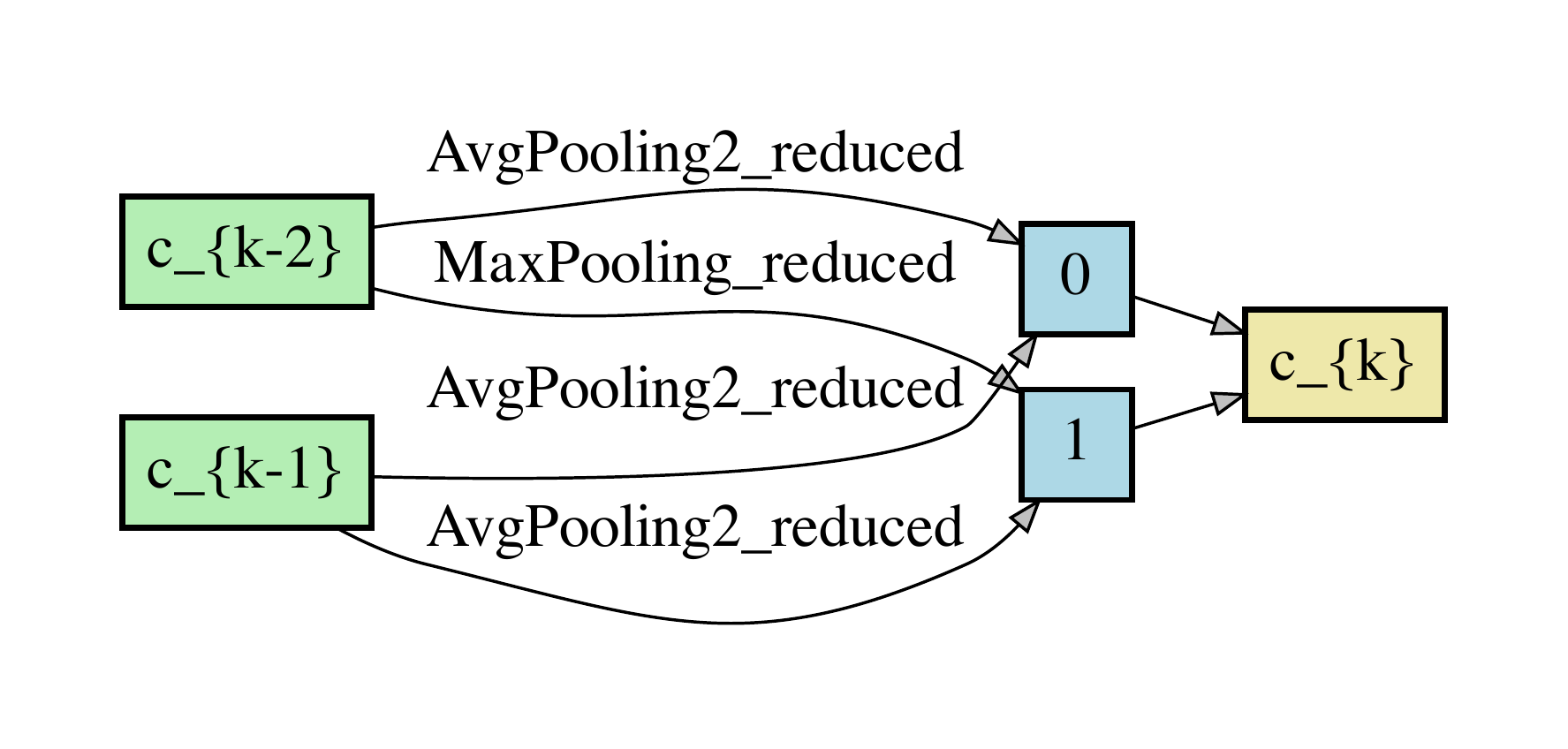}}
\caption{\small (a)-(b) Normal cell and Reduction cell for multiple dimensionality reduction respectively} \label{fig:dimension_reduce}
\end{figure*}

To study the behaviour of our method under multiple dimesionality reduction pipeline on HDM05, we use the preprocessing layers to perform dimensionality reduction. To be precise, we consider a \textbf{preprocessing} step to reduce the dimension from 93 to 70 to 50 and then, a \textbf{reduction cell} that reduced the dimension from 50 to 24. This modification has the advantage that it reduces the search time from 3 CPU days to 2.5 CPU days, and in addition, provides a performance gain (see Table (\ref{tab:multidimred})). The normal and the reduction cells for the multiple dimension reduction are shown in Figure (\ref{fig:dimension_reduce}).

\begin{table}[h]
\scriptsize
\centering
\begin{tabular}{cccccc}
\hline
\rowcolor[gray]{0.85}
 Preprocess dim reduction        &  Cell dim reduction  & SPDNetNAS & Search time   \\ \hline
 NA & 93 $\rightarrow$ 30 & 68.74\% $\pm$ 0.93 & 3 CPU days \\ \hline
 93$\rightarrow$70$\rightarrow$50 & 50$\rightarrow$24  & \textbf{69.41 \% $\pm$ 0.13} & 2.5 CPU days \\ \hline
\end{tabular}
\caption{ \small{Results of modifying preprocessing layers for multiple dimentionality reduction on HDM05}}\label{tab:multidimred}
\end{table}

\subsection{Effect of adding nodes to the cell}
\begin{figure*}[!htbp]
\centering
    \subfigure[\label{fig:multinodenormal} Normal cell]{\includegraphics[width=0.5\linewidth]{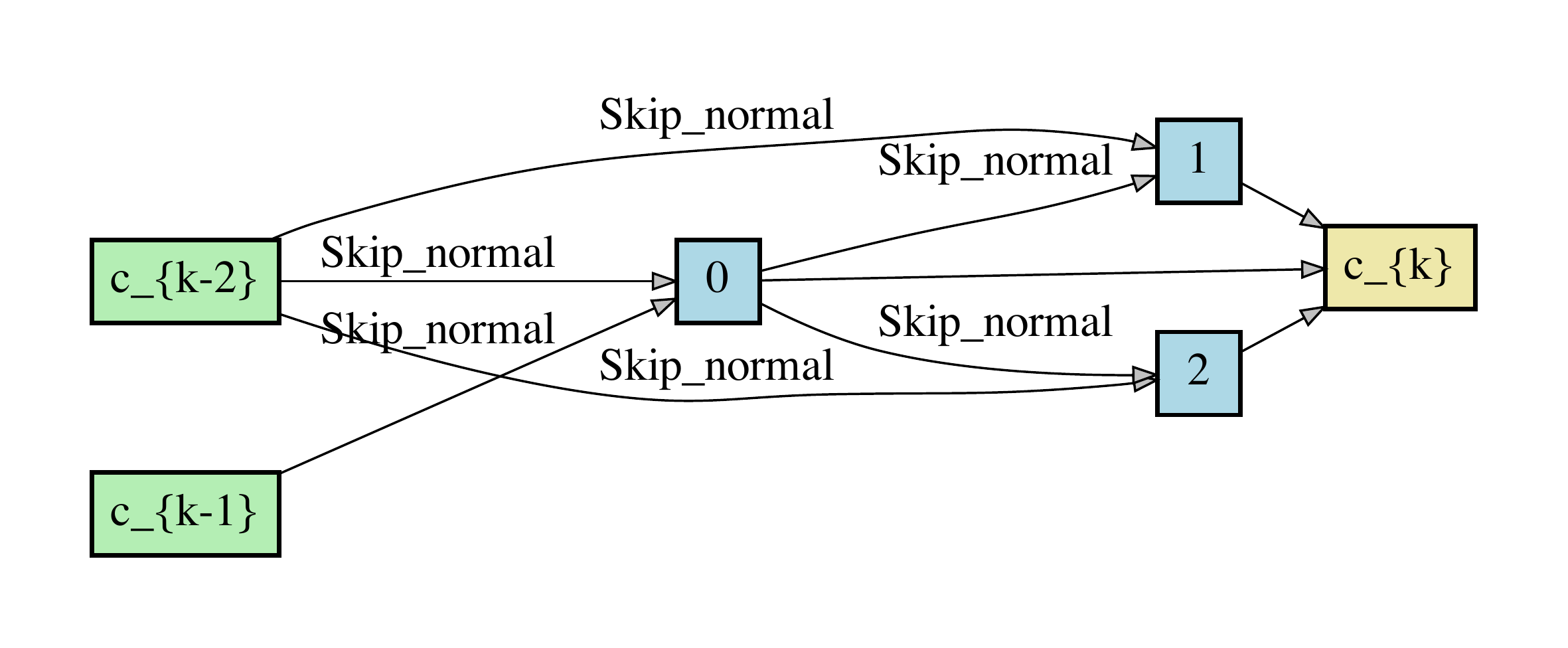}}
    \subfigure[\label{fig:multinodereduce} Reduction cell]{\includegraphics[width=0.38\linewidth]{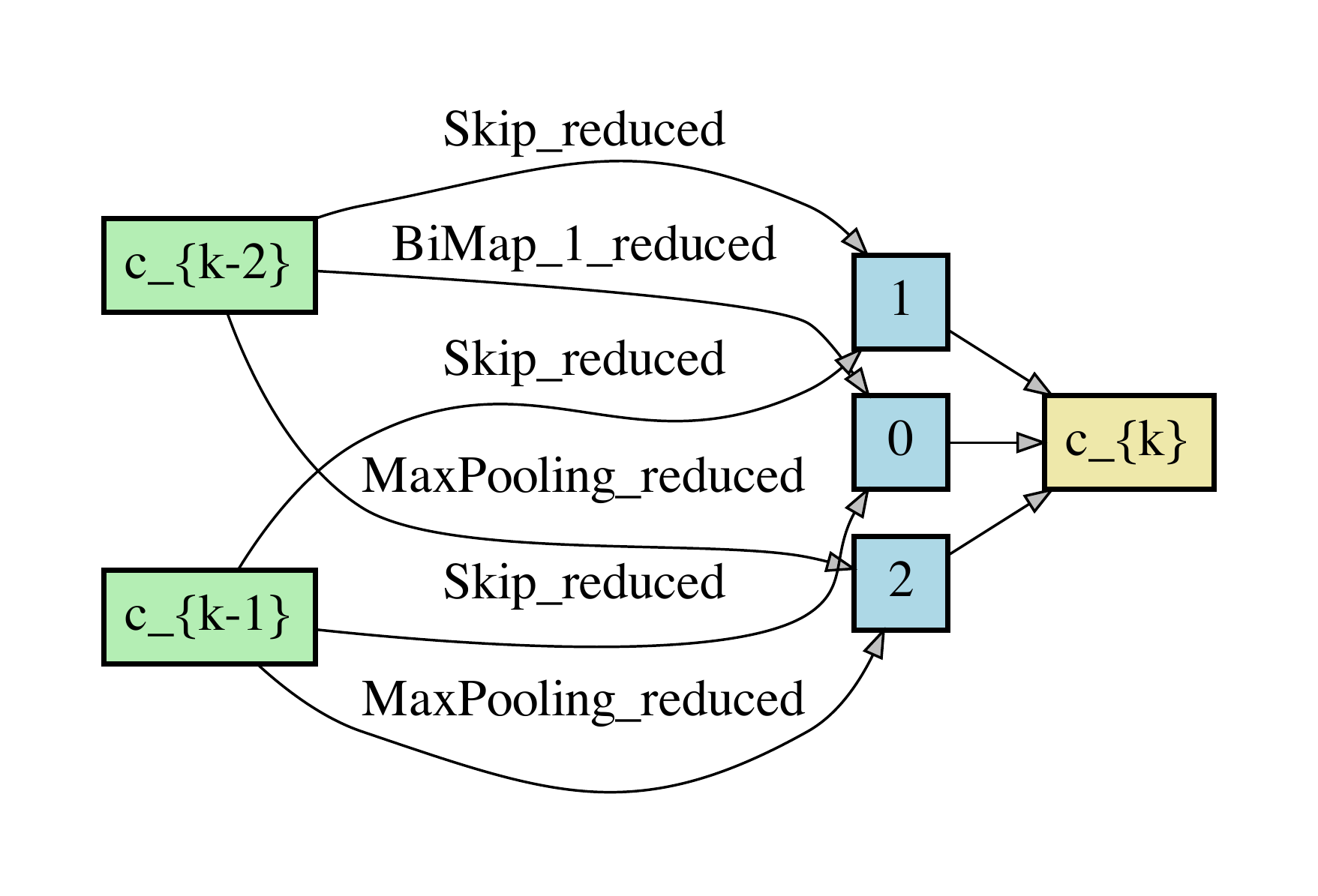}}
\caption{\small (a)-(b)  Optimal Normal cell and Reduction cell with 6 nodes on the HDM05 dataset}\label{fig:extranode}
\end{figure*}

Experiments presented in the main paper consists of $N = 5$ nodes per cell which includes two input nodes, one output node, and two intermediate nodes. To do further analysis of our design choice, we added nodes to the cell. Such analysis can help us study the critical behaviour of our cell design i.e, whether adding an intermediate nodes can improve the performance or not?, and how it affects the computational complexity of our algorithm? To perform this experimental analysis, we used HDM05 dataset \cite{muller2007documentation}. We added one extra intermediate node ($ N = 6 $) to the cell design. We observe
that we converge towards an architecture design that is very much similar in terms of operations (see Figure \ref{fig:extranode}). The evaluation results shown in Table (\ref{tab:morenodes}) help us to deduce that adding more intermediate nodes increases the number of channels for output node, subsequently leading to increased complexity and almost double the computation time.

\begin{table}[h]
\scriptsize
\centering
\begin{tabular}{cccccc}
\hline
\rowcolor[gray]{0.85}
Number of nodes  & SPDNetNAS & Search time \\ \hline
5  & \textbf{68.74\% $\pm$ 0.93} & 3 CPU days \\ \hline
6  & 67.96\% $\pm$ 0.67 & 6 CPU days \\ \hline
\end{tabular}
\caption{ \small{Results for multi-node experiments on HDM05}}\label{tab:morenodes}
\end{table}

\subsection{Effect of adding  multiple cells}
In our paper we stack 1 normal cell over 1 reduction cell for all the experiments. For more extensive analysis of the proposed method, we conducted training experiments by stacking multiple cells which is in-line with the experiments conducted by \cite{liu2018darts}. We then transfer the optimized architectures from the singe cell search directly to the multi-cell architectures for training. Hence, the search time for all our experiments is same as for a single cell search i.e. 3 CPU days. Results for this experiment are provided in Table \ref{tab:multicell}. The first row in the table shows the performance for single cell model, while the second and third rows show the performance with multi-cell stacking. \textit{Remarkably, by stacking multiple cells our proposed SPDNetNAS outperforms SPDNetBN \cite{brooks2019riemannian} by a large margin (about \textbf{8\%}, i.e., about \textbf{12\%} for the relative improvement)}.

\begin{table*}[h!]
\scriptsize
\centering
\begin{tabular}{c|c|c|c|cc}
\hline
\rowcolor[gray]{0.85}
 & Dim reduction in cells & Cell type sequence & SPDNetNAS & Search Time \\ \hline
single cell & $93\rightarrow46$ & reduction-normal & 68.74\% $\pm$ 0.93 & 3 CPU days\\ \hline
multi-cell & $93\rightarrow46$ & normal-reduction-normal  & \textbf{71.48\% $\pm$ 0.42} & 3 CPU days \\ \hline
multi-cell & $93\rightarrow46\rightarrow22$ & reduction-normal-reduction-normal  & \textbf{73.59 \% $\pm$ 0.33} & 3 CPU days\\
  \hline
\end{tabular}
\caption{\small{Results for multiple cell search and training experiments on HDM05: reduction corresponds to reduction cell and normal corresponds to the normal cell.}}\label{tab:multicell}
\end{table*}

\subsection{AFEW Performance Comparison on RAW SPD features}

In addition to the evaluation on CNN features in the major paper, we also use the raw SPD features (extracted from gray video frames) from \cite{huang2017riemannian,brooks2019riemannian} to compare the competing methods. To be specific, each frame is normalized to $20 \times 20$ and then represent each video using a $400 \times 400$ covariance matrix \cite{wang2012covariance,huang2017riemannian}. Table~\ref{tab:comparison_afew_raw} summarizes the results. As we can see, the transferred architecture can handle the new dataset quite convincingly. The test accuracy is comparable to the best SPD network method for RADAR model transfer. For HDM05 model transfer, the test accuracy is much better than the existing SPD networks.

\begin{table*}[h]
\scriptsize
\centering
\begin{tabular}{cc|ccc|ccc}
\hline
\rowcolor[gray]{0.85}
DARTS & FairDARTS & ManifoldNet & SPDNet  & SPDNetBN  & Ours(R) &  Ours(RADAR) &  Ours (HDM05) \\ \hline
25.87 \% & 25.34\% & 23.98\% & 33.17\%  & 35.22\%  & 32.88\% & \textbf{ 35.31} \%  & \textbf{38.01\%} \\ \hline
\end{tabular}
\caption{\small Performance of transferred SPDNetNAS Network architecture in comparison to existing SPD Networks on the AFEW dataset. RAND symbolizes random architecture from our search space. DARTS/FairDARTS: accepts the logarithms of raw SPDs, and the other competing methods receive the SPD features.}\label{tab:comparison_afew_raw}
\end{table*}

\subsection{Derived cell architecture using sigmoid on Fr\'echet mixture of SPD operation}
\begin{figure*}[t]
\centering
    \subfigure[\label{fig:RADAR_HDM_genotypes_softmax}Derived cell architecture using softmax]{\includegraphics[width=0.48\linewidth]{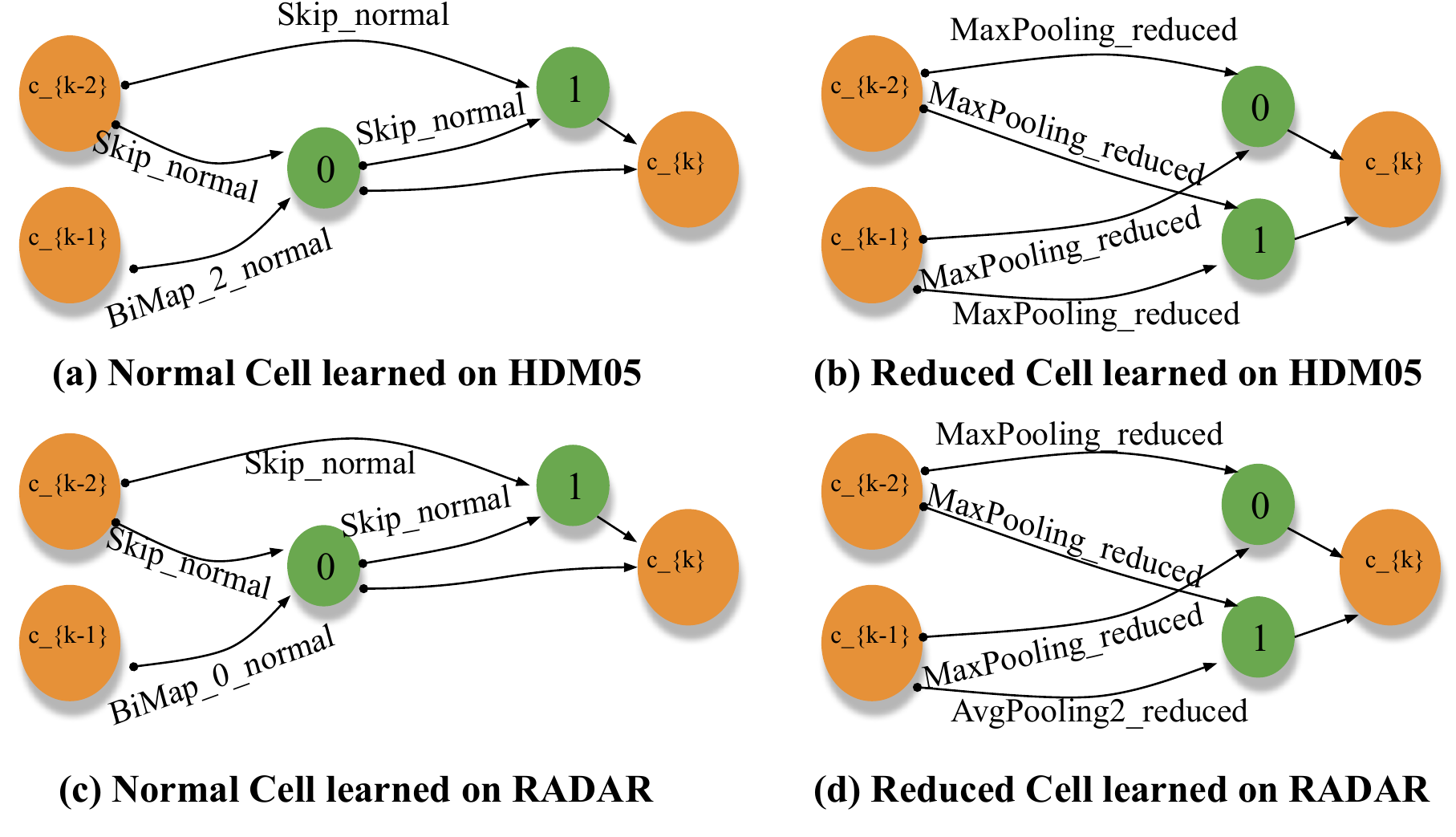}}
    \subfigure[\label{fig:RADAR_HDM_genotypes_sigmoid}Derived cell architecture using sigmoid]{\includegraphics[width=0.48\linewidth]{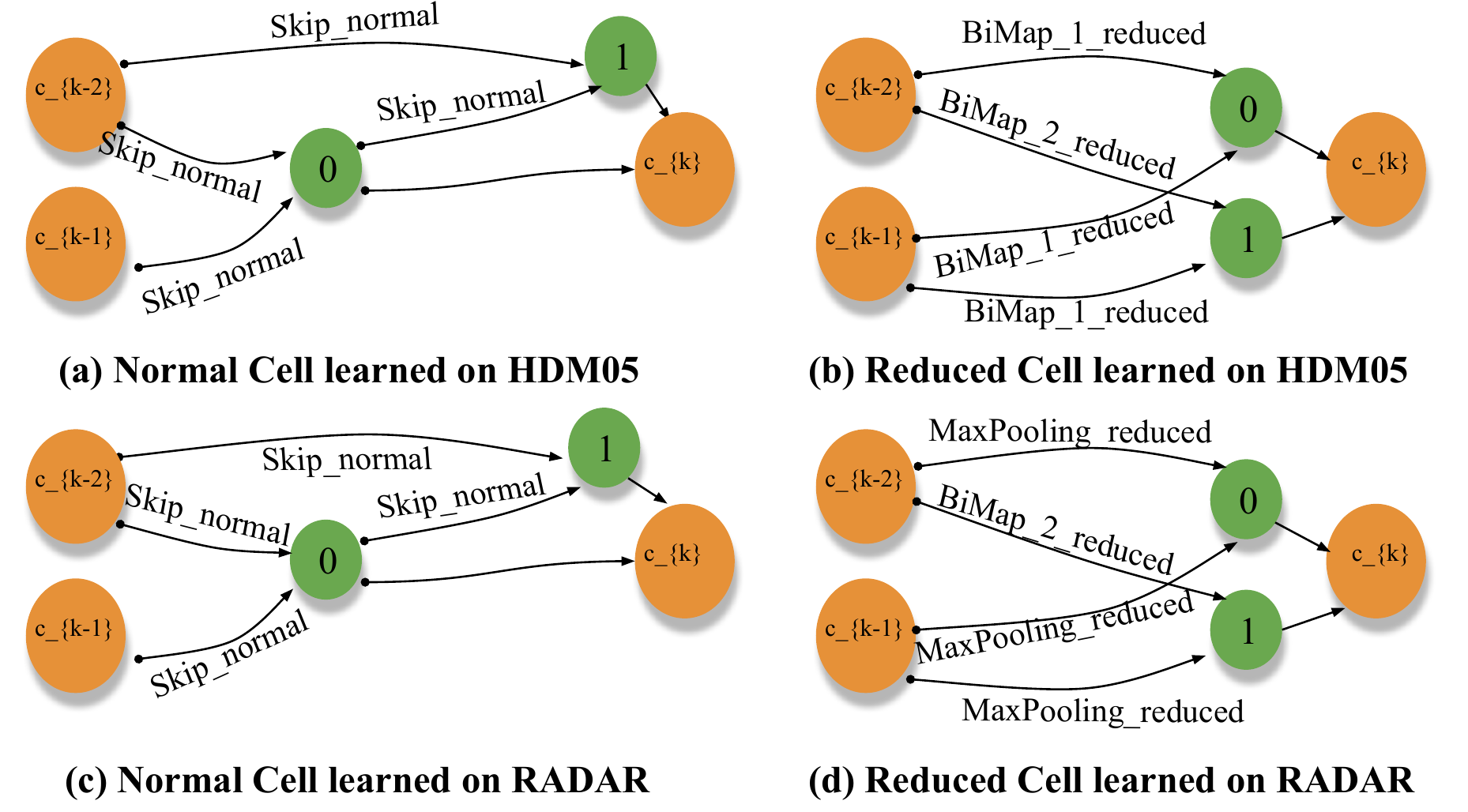}}
\caption{\small(a), (b) Derived architecture by using softmax and sigmoid on the Fr\'echet mixture of SPD operations. These are the normal cell and reduced cell obtained on RADAR and HDM05 dataset. }
\label{fig:genotype_softmax_sigmoid}
\end{figure*}

Figure \ref{fig:RADAR_HDM_genotypes_softmax} and Figure \ref{fig:RADAR_HDM_genotypes_sigmoid} show the cell architecture obtained using the softmax and sigmoid respectively on the  Fr\'echet mixture of SPD operation. It can be observed that it has relatively more skip and pooling operation than sparsemax. In contrast to softmax and sigmoid, the SPD cell obtained using sparsemax is composed of more convolution type operation in the architecture, which in fact is important for better representation of the data.  

\subsection{Comparison between Karcher flow and recursive apporach for weighted Frechet mean}


The proposed NAS algorithm is based on Fréchet Mean computations. From the weighted mixture of operations between nodes to the derivation of intermediate nodes, both compute the Fréchet mean of a set of points on the SPD manifold. It is well known that there is no closed form solution when the number of input samples is bigger than 2 \cite{brooks2019riemannian}. We can only compute an approximation using the famous Karcher flow algorithm \cite{brooks2019riemannian} or recursive geodesic mean \cite{chakraborty2020manifoldnet}. For comparison, we replace our used Karcher flow algorithm with the recursive approach under our SPDNetNAS framework. Table \ref{tab:frechet} sumarizes the comparison between these two algorithms. 
We observe considerable decrease in accuracy for both the training and test set when using the recursive methods, showing that the Karcher flow algorithm favors our proposed algorithm more.

\begin{table}[!htbp]
\scriptsize
\centering
\begin{tabular}{cccc}
\hline
\rowcolor[gray]{0.85}
Dataset / Method & Karcher flow  & Recursive algorithm  \\ \hline
RADAR & \textbf{96.47\% $\pm$ 0.08}  & 68.13\% $\pm$ 0.64  \\  \hline
HDM05 & \textbf{68.74\% $\pm$ 0.93}  & 56.85\% $\pm$ 0.17  \\ \hline\\
\end{tabular}
\caption{ \small{Test performance of the proposed SPDNetNAS using the Karcher flow algorithm  and the recursive algorithm to compute Fréchet means.}}\label{tab:comparisonfrechetsolvers}
\label{tab:frechet}
\vspace{-1em}
\end{table}

\subsection{Convergence curve analysis}

\begin{figure}[t]
\centering
    \subfigure[\label{fig:RADAR_ValVsEpoch}]{\includegraphics[width=0.45\linewidth]{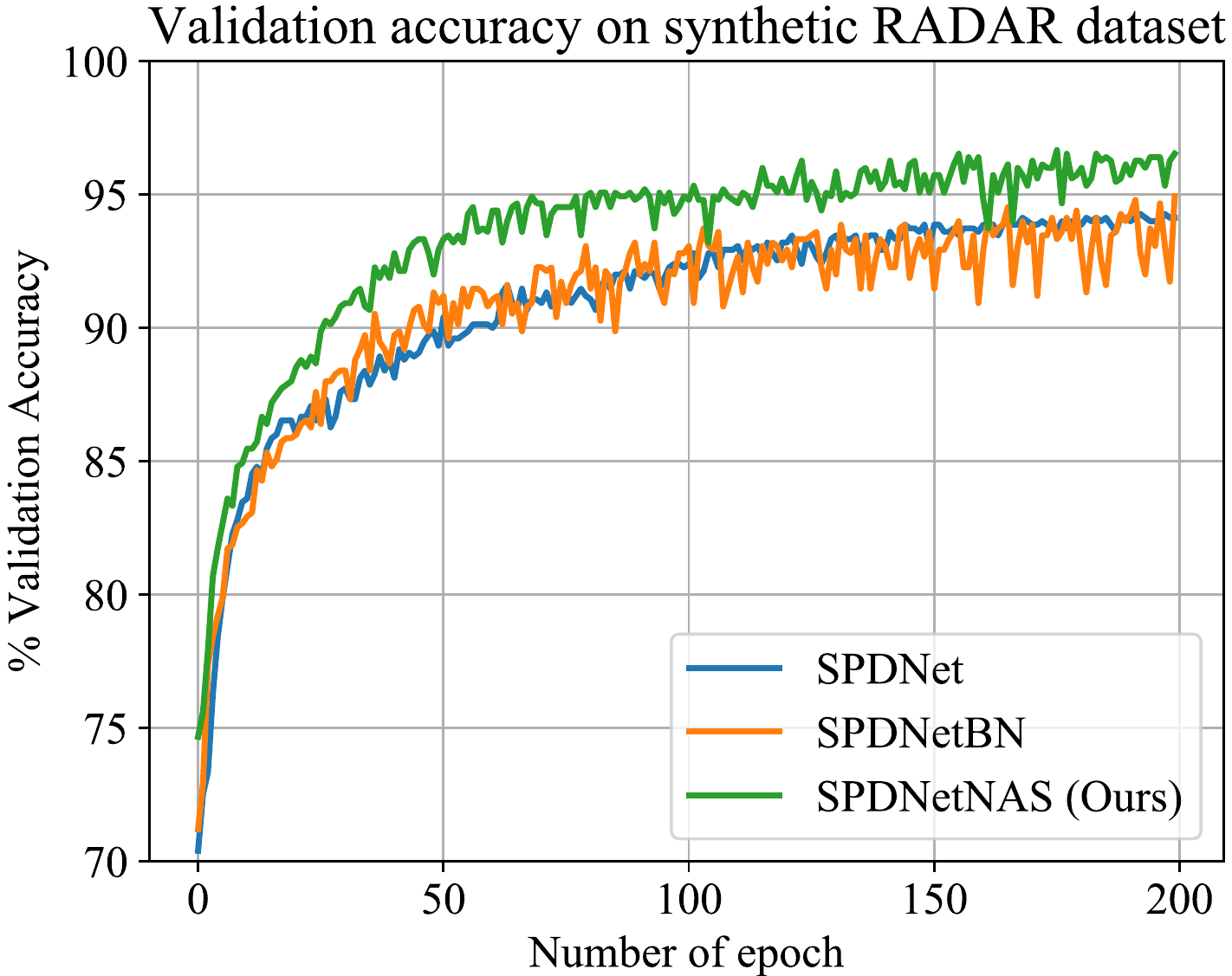}}
    \subfigure[\label{fig:RADAR_TestVsSample}]{\includegraphics[width=0.45\linewidth]{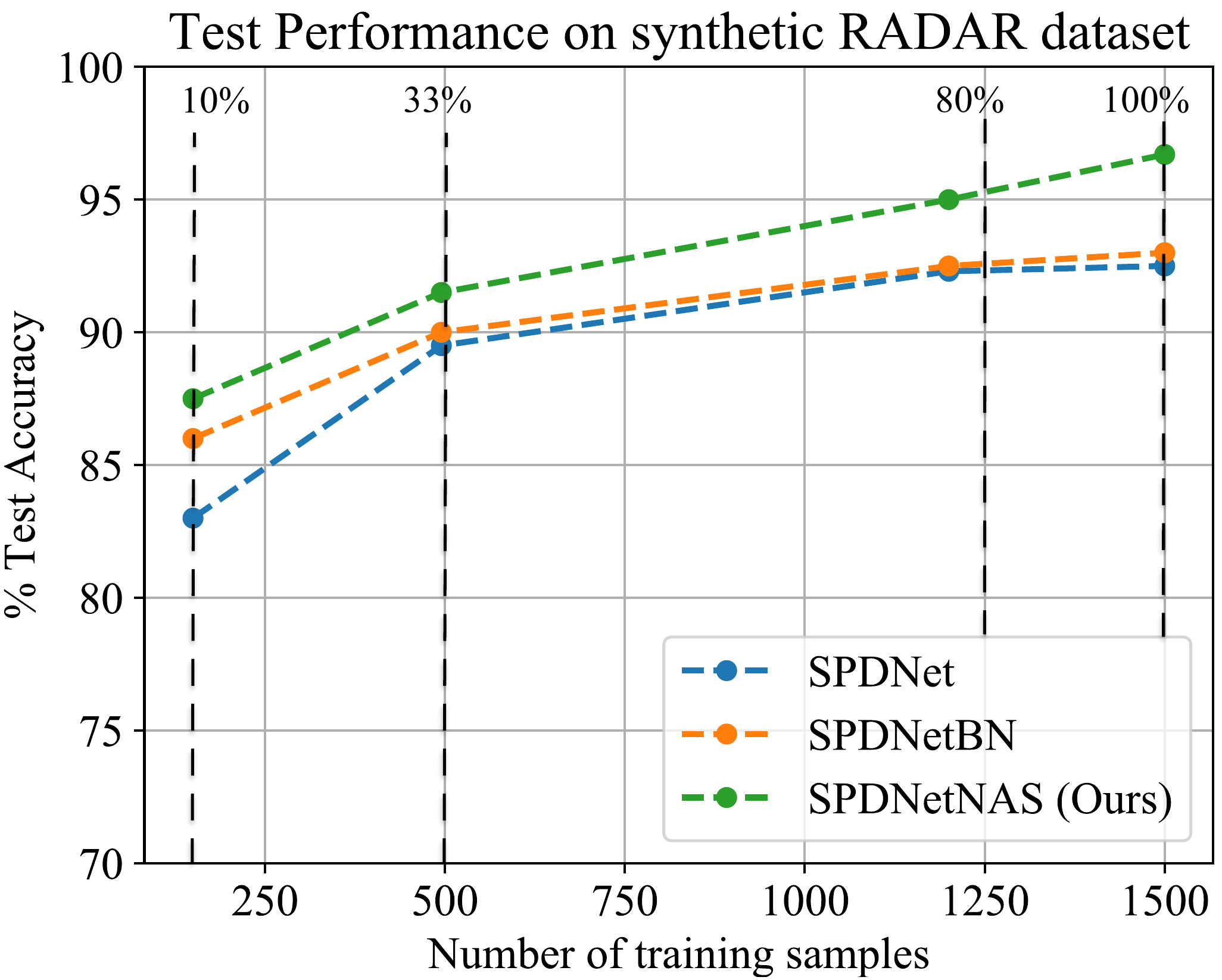}}
\caption{\small (a)  Validation accuracy of our method in comparison to the SPDNet  and SPDNetBN  on RADAR dataset. Clearly, our SPDNetNAS algorithm show a steeper validation accuracy curve. (b) Test accuracy on 10\%, 33\%, 80\%, 100\% of the total data sample. It can be observed that our method exhibit superior performance.}
\label{fig:stats_appendix}
\end{figure}

Figure \ref{fig:RADAR_ValVsEpoch}  shows the validation curve which almost saturates at 200 epoch demonstrating the stability of our training process. First column bar of Figure~(\ref{fig:RADAR_TestVsSample}) show the test accuracy comparison when only 10\% of the data is used for training our architecture which demonstrate the effectiveness of our algorithm. Further, we study this for our SPDNetNAS architecture by taking 10\%, 33\%, 80\% of the data for training. Figure \ref{fig:RADAR_TestVsSample}) clealy show our superiority of SPDNetNAS algorithm than handcrafted SPD networks.

\begin{figure}[!htbp]
\centering
    \subfigure[\label{fig:Radarloss}]{\includegraphics[width=0.45\linewidth]{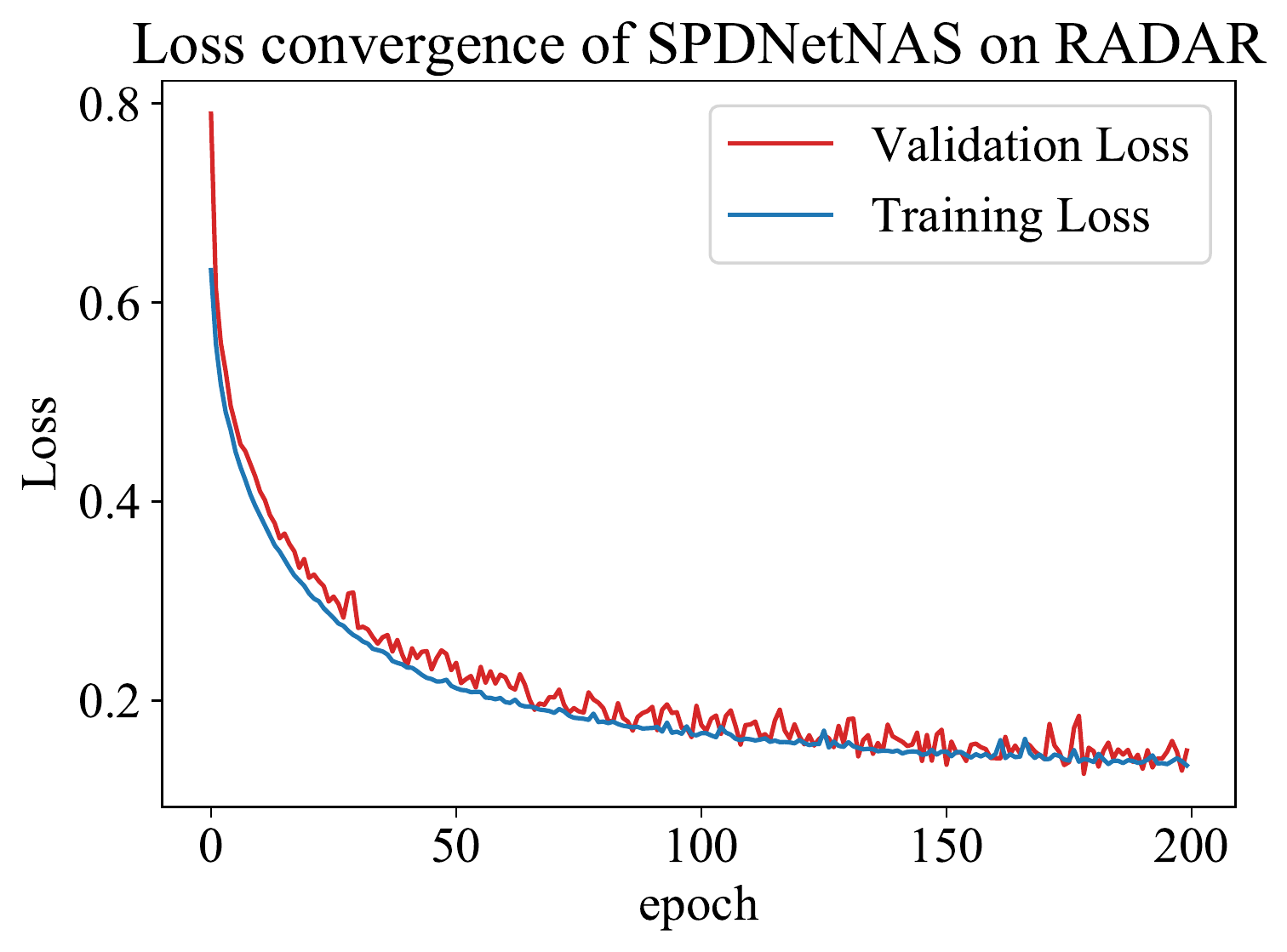}}
    \subfigure[\label{fig:HDMloss}]{\includegraphics[width=0.45\linewidth]{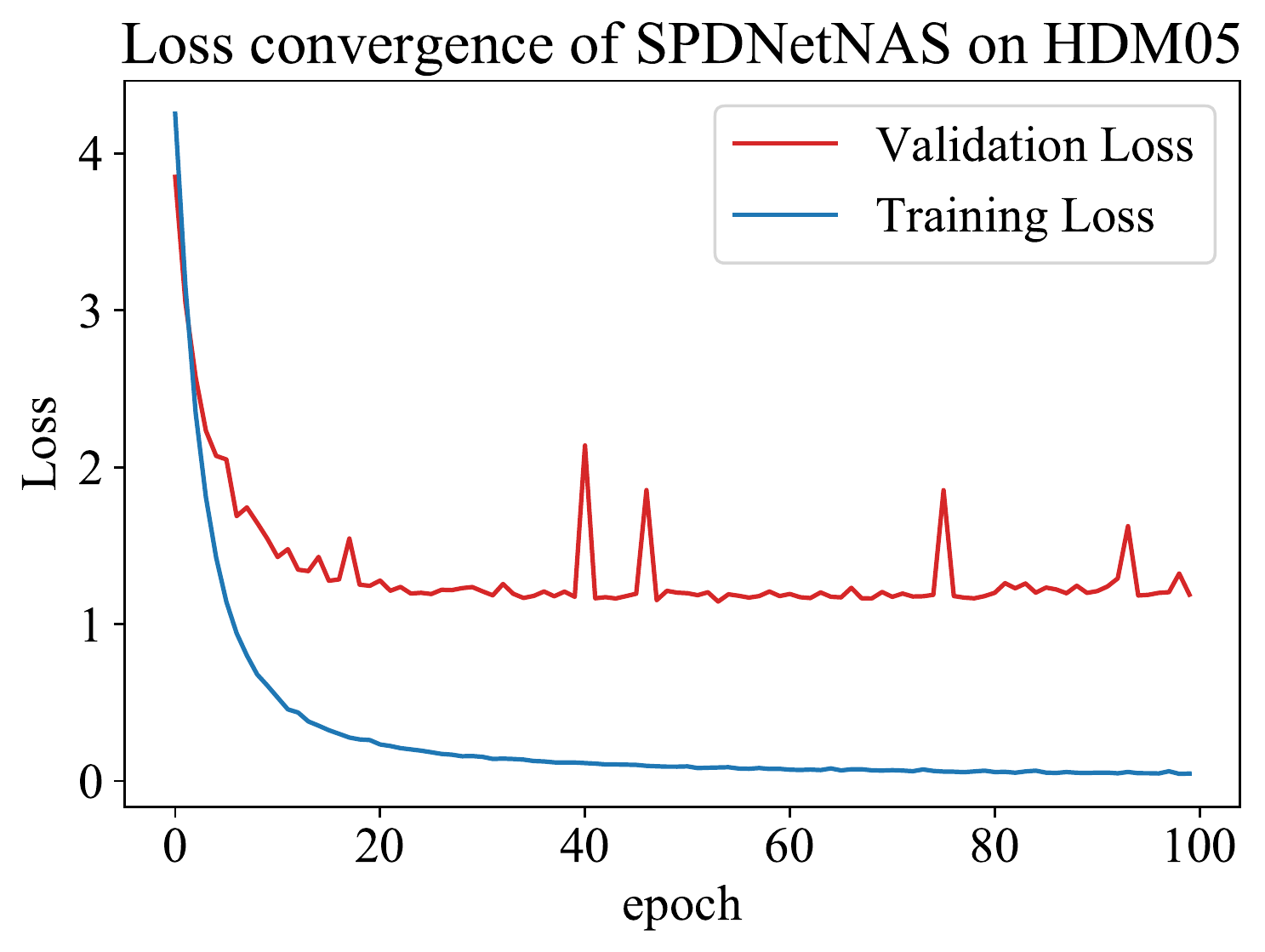}}
\caption{\small (a) Loss function curve showing the values over 200 epochs for the RADAR dataset (b) Loss function curve showing the values over 100 epochs on the HDM05 dataset. }\label{fig:lossfct}
\end{figure}

Figure (\ref{fig:Radarloss}) and Figure (\ref{fig:HDMloss}) show the convergence curve of our loss function on the RADAR and HDM05 datasets respectively. For the RADAR dataset the validation and training losses follow a similar trend and converges at 200 epochs. For the HDM05 dataset, we observe the training curve plateaus after 60 epochs, where as  the validation curve takes 100 epochs to provide a stable performance. Additionally, we noticed a reasonable gap between the training loss and validation loss for the HDM05 dataset \cite{muller2007documentation}. A similar pattern of convergence gap between validation loss and training loss has been observed by  \cite{huang2017riemannian} work.

\subsection{Why we preferred to simulate our experiments on CPU rather than GPU?}

When dealing with SPD matrices, we need to carry out complex computations. These computations are performed to make sure that our transformed representation and corresponding operations respect the underlying manifold structure. In our study, we analyzed SPD matrices with the Affine Invariant Riemannian Metric (AIRM), this induces operations heavily dependent on singular value decomposition (SVD) or eigendecomposition (EIG).
Both decompositions suffer from weak support on GPU platforms. Hence, our training did not benefit from GPU acceleration and we decided to train on CPU. As a future work, we aim to speedup our implementation on GPU by optimizing the SVD Householder bi-diagonalization process as studied in some existing works like \cite{dong2017optimizing,gates2018accelerating}.
	
\end{document}